\documentclass{isprs} 
\usepackage{fix-cm}
\usepackage{setspace}
\usepackage{geometry} 
\usepackage{epstopdf}
\usepackage[labelsep=period]{caption}  
\usepackage[british]{babel} 
\usepackage[hang]{footmisc}
\usepackage{amsmath,amssymb} 
\usepackage{booktabs}
\usepackage{subcaption}
\usepackage{float}
\usepackage{booktabs} 
\usepackage{multirow} 
\usepackage{siunitx}  
\usepackage{xcolor}   
\usepackage{colortbl}  

\title{Structured-Li-GS: Structured 3D Gaussians Splatting with LiDAR Incorporation and Spatial Constraints}
\date{}

\begin{document}


\author{\centering
    Huaiyuan Weng\textsuperscript{1},
    Huibin Li\textsuperscript{1},
    Chul Min Yeum\textsuperscript{1}
}

\abstract{
In this study, we develop a \underline{Structured} framework for \underline{G}aussian \underline{S}platting (3DGS) with \underline{Li}DAR integration (Structured-Li-GS). It is a lightweight Gaussian Splatting pipeline that leverages LiDAR–inertial–visual SLAM. Structured-Li-GS achieves high-quality 3D reconstructions with fewer Gaussians by training on accurate, dense, colorized point clouds. Gaussian primitives are anchored using sub-sampled point clouds, and their ellipsoidal parameters are initialized from local surface geometry. Our training strategy integrates a comprehensive set of loss terms, including photometric, flattening, offset, depth, and normal losses—guided by the dense point cloud, enabling accurate reconstruction without Gaussian densification. This approach produces up-to-scale, high-fidelity results with a moderate model size. For experimental validation, we develop a custom hardware-synchronized LiDAR–camera handheld scanner. Experiments on both benchmark datasets and our real-world in-house dataset demonstrate that Structured-Li-GS surpasses state-of-the-art methods while using fewer Gaussians.
}


\address{
  \textsuperscript{1} Department of Civil and Environmental Engineering, University of Waterloo, 200 University Ave W, Waterloo, N2L 3G1, \\ Ontario, Canada - (huaiyuan.weng, matt.li, cmyeum)@uwaterloo.ca\\
}
\maketitle


\section{Introduction}

The evolution of 3D reconstruction techniques has been marked by significant milestones in robotics. Photogrammetry reconstruction (e.g. SfM (Structure-from-Motion)) \cite{schonberger2016structure}, and SLAM (Simultaneous Localization and Mapping) \cite{zhang2014loam,shan2020lio,zheng2022fast,zheng2024fast}, laid the foundation by generating point clouds, voxel grids, or mesh-based map reconstructions. However, these representations of maps often lack photorealistic detail and are not well-suited for high-quality rendering. In recent years, researchers have explored volume rendering, which can be broadly categorized into neural implicit and explicit representations. Neural implicit representations, such as Neural Radiance Fields (NeRF) \cite{mildenhall2021nerf} and Neural Signed Distance Functions \cite{park2019deepsdf}, encode 3D scenes using continuous functions, enabling high-quality rendering but often at the cost of computational efficiency. On the other hand, explicit representations, such as 3D Gaussian Splatting (3DGS) \cite{kerbl3Dgaussians}, represent scenes using discrete 3D Gaussian ellipsoids, offering a significant reduction in training time and enabling real-time rendering.

Despite the progress of 3DGS, achieving geometrically accurate reconstructions with moderate model sizes remains challenging. Several factors contribute to this limitation. First, 3DGS struggles to accurately represent thin or surface-like structures because the ellipsoidal Gaussian assumption inadequately models such geometry, especially in outdoor or unbounded environments.
Second, 3DGS typically relies on photometric loss with sparse supervision from SfM point clouds, resulting in limited geometric constraints and suboptimal reconstruction accuracy. Finally, densification procedures used in existing methods introduce a large number of Gaussians, increasing model size and prolonging training and inference, particularly for large-scale scenes.

\begin{figure}[t]
    \centering
    \includegraphics[width=0.48\textwidth]{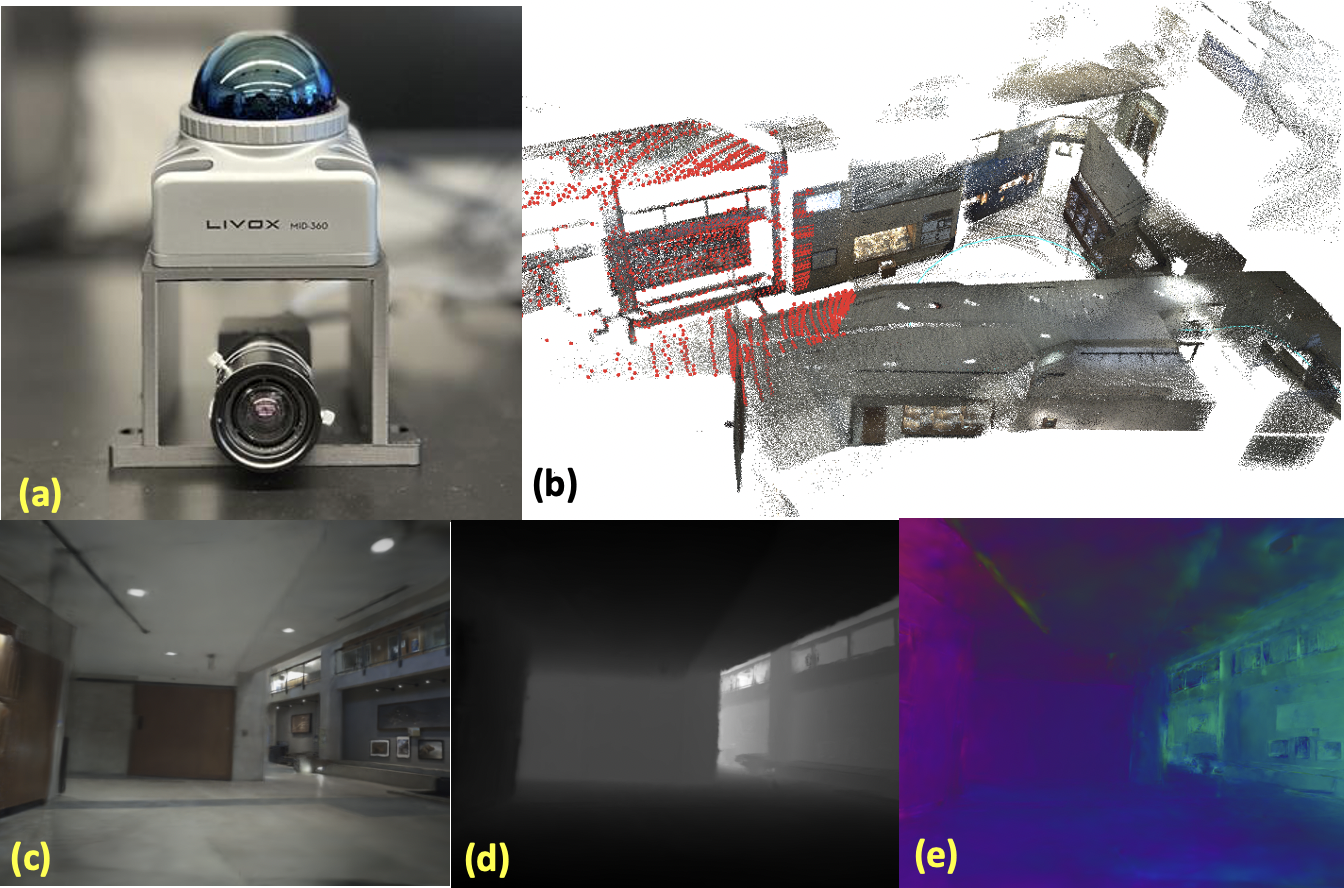}
    \caption{The performance of Structured-Li-GS. (a) Our custom-designed, hardware-synchronized handheld scanner equipped with LiDAR and camera sensors. (b) A colorized 3D point cloud generating using LiDAR-inertial-RGB SLAM and data collected from the handheld scanner. (c)–(e): Rendered outputs of an indoor scene: (c) RGB image, (d) depth map, and (e) surface normal map.}
    \label{teaser}
\end{figure}

LiDAR–visual fusion offers a promising path forward. LiDAR provides accurate depth measurements over long ranges, and when combined with visual and inertial cues through modern SLAM systems, yields precise poses and high-quality dense point clouds. These multi-modal priors can significantly benefit 3DGS reconstruction, yet the integration of LiDAR-SLAM data into 3DGS training remains largely unexplored.

In this study, as shown by Fig.~\ref{teaser}, we propose Structured-Li-GS, a structured, light-aware 3DGS reconstruction framework that integrates LiDAR with Gaussian Splatting to introduce strong geometric constraints, thereby improving reconstruction accuracy while maintaining a compact model size. Structured-Li-GS produces high-fidelity 3D reconstructions using  fewer Gaussians without compromising their quality by leveraging dense, colorized LiDAR point clouds: Gaussians are anchored on sub-sampled LiDAR points, and their ellipsoids are initialized using local surface geometry. The model is trained using a comprehensive combination of photometric, geometric, and depth losses, enabling accurate reconstruction without the need for densification.

The key contributions of this work are as follows:
\begin{itemize}
\item A robust data-preparation pipeline that integrates a LiDAR inertial visual odometry algorithm with advanced post processing to generate a high-quality dataset tailored for 3DGS training.
\item A normal-assisted initialization strategy for anchored Gaussian ellipsoids that improves geometric accuracy, enhances surface fitting, and reduces model size.
\item A comprehensive loss formulation—incorporating RGB, flatten, offset, depth, and normal losses—to jointly optimize photometric and geometric consistency.
\item Real-world evaluation using a custom LiDAR–camera handheld scanner, demonstrating both qualitative and quantitative improvements.
\end{itemize}

\begin{figure*}[t] 
    \centering
    \includegraphics[width=0.98\textwidth]{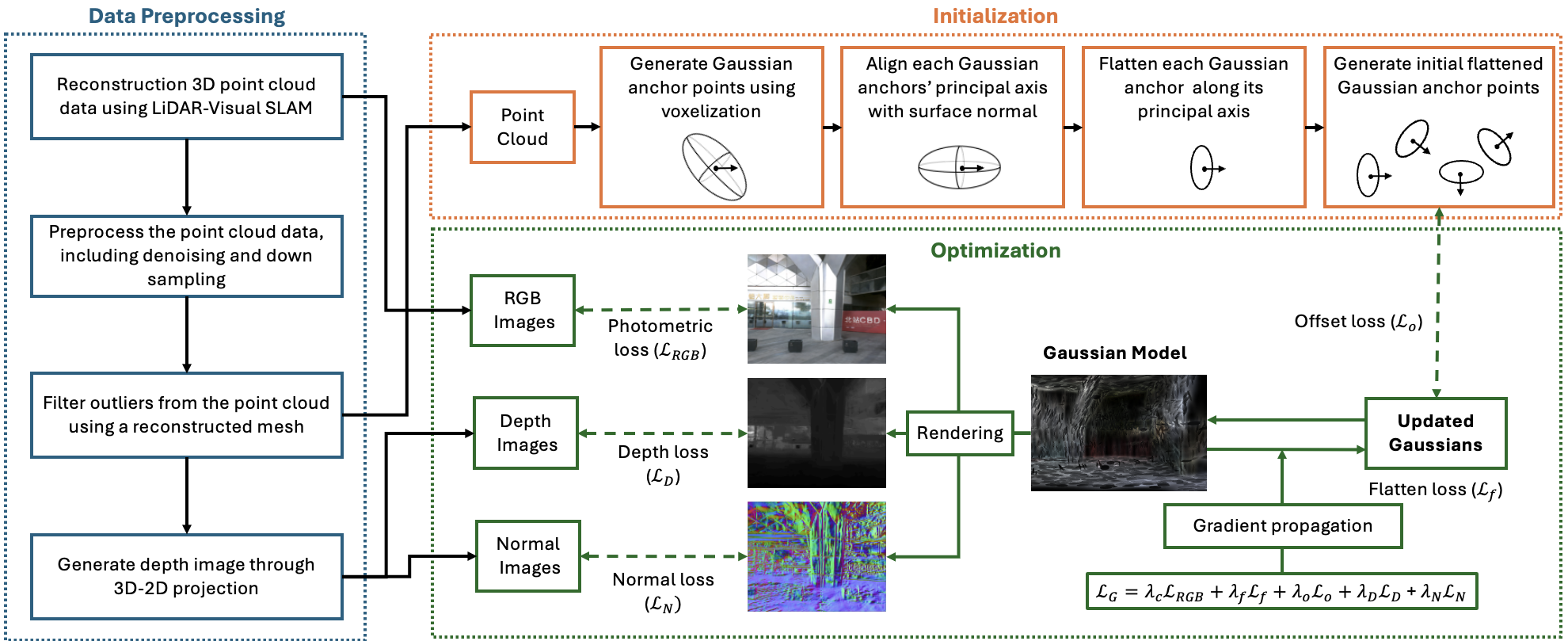}
    \caption{Structured-Li-GS system architecture overview. Our system involves three steps: (a) We begin with data preprocessing using LiDAR-Visual SLAM to generate and filter point clouds, (2) followed by Gaussian anchor initialization through voxelization and alignment with surface normals. (3) Finally, the Gaussian model is optimized via rendering losses from RGB, depth, and normal images to iteratively refine the scene representation.}
    \label{fig:big_diagram}
\end{figure*}

\section{Related Work}
Integrating LiDAR data can enhance 3DGS in large-scale map reconstruction. LiDAR-based 3DGS methods can be categorized into two main approaches. The first approach directly integrates 3DGS into the SLAM process, where image sequences are used for both pose tracking and reconstruction. For example, MM-Gaussian \cite{wu2024mm} utilizes a point cloud-Gaussian registration algorithm to estimate camera poses and integrates these point clouds directly into the map. LiV-GS \cite{xiao2024liv}, a LiDAR-visual SLAM system, adopts 3D Gaussian as a differentiable spatial representation, enabling direct alignment of sparse LiDAR data with continuous Gaussian maps. However, this direct LiDAR-visual SLAM approach to 3DGS often struggles with accurate 3D reconstruction due to challenges in precise pose estimation. Unlike traditional point clouds, Gaussian maps represent surfaces probabilistically, introducing ambiguity in feature matching and alignment and leading to inconsistent reconstruction accuracy.

The second approach utilizes point cloud maps and poses derived from LiDAR SLAM as initial inputs, replacing conventional SfM data. Gaussian-LIC \cite{lang2024gaussian}, for instance, constructs 3D maps by directly integrating LiDAR point clouds and images, resulting in a more detailed environmental representation. LetsGo \cite{letsgo} introduces Level of Detail (LoD) rendering in 3DGS, employing multi-resolution Gaussian functions to represent 3D scenes while using a hierarchical map structure for efficient storage and rendering of large-scale environments. LIV-GaussianMap \cite{hong2024liv} employs an explicit octree structure for point cloud management and optimizes Gaussians by differentiable ellipsoidal surface representations. LiDAR-3DGS \cite{lim2024lidar} preprocesses LiDAR data from LiDAR-inertial-visual SLAM lacks further refinement in the 3DGS algorithm. LI-GS \cite{jiang2024li} proposes an incremental multimodal 3DGS reconstruction method that incorporates plane constraints using Gaussian Mixture Models,  enhancing geometric accuracy, particularly for mesh-based models. 

Despite these advancements in LiDAR-based 3DGS, most techniques largely enhance reconstruction accuracy by increasing the input point cloud density and densifying Gaussians while training. However, these approaches often lead to a substantial increase in the number of Gaussians, resulting in a large and unwieldy 3D Gaussian model. This leads to challenges in scalability and efficiency, making it difficult to apply these methods to large-scale environments.

\section{Methodology}
An overview of Structured-Li-GS is presented in Fig.~\ref{fig:big_diagram}. This section introduces the key components and their process: Data Preprocessing (\ref{subsec:data_preprocess}), Initialization (\ref{subsec:initialization}), and Optimization (\ref{subsec:optimization}).

\subsection{Preliminaries}
\label{subsec:preliminaries}
3DGS represents the scene with a set of anisotropic
3D Gaussians \cite{kerbl3Dgaussians}. This approach inherits the differential properties of volumetric representation while achieving efficient rendering through tile-based rasterization.

In vanilla 3DGS, the process begins with a set of 3D points reconstructed from SfM, where each point corresponds to the position (mean) $\mu$ of a 3D Gaussian. The 3D Gaussian function, $G(x)$,  is defined as
\[
G(x) = e^{-\frac{1}{2}(x - \mu)^{T}\Sigma^{-1}(x - \mu)},
\]
where \(x\) is an arbitrary position within the 3D scene and \(\Sigma\) denotes the covariance matrix of the 3D Gaussian. To ensure the positive semi-definite property of \(\Sigma\), it is formulated using a scaling matrix \(S\) and a rotation matrix \(R\) as follows:
\[
\Sigma = RSS^{T}R^{T}.
\]

In addition to color \(c\) modeled by spherical harmonics, each 3D Gaussian is associated with an opacity \(\alpha\) which is multiplied by \(G(x)\) during the blending process.

Unlike conventional volumetric rendering methods, 3DGS utilizes a tile-based rasterization technique instead of computationally expensive ray-marching. \(G(x)\) is first transformed to 2D Gaussians \(G^{\prime}(x)\) on the image plane following the projection process \cite{kerbl3Dgaussians}. A tile-based rasterizer then efficiently sorts the 2D Gaussians and applies \(\alpha\)-blending using the equation: 
\[
C(x^{\prime}) = \sum_{i \in N} c_{i} \sigma_{i} \prod_{j=1}^{i-1} (1 - \sigma_{j}), \quad \sigma_{i} = \alpha_{i} G^{\prime}_{i}(x^{\prime}),
\]
where \(x^{\prime}\) is the queried pixel position and \(N\) denotes the number of sorted 2D Gaussians associated with the queried pixel. By leveraging the differentiable rasterizer, all attributes of the 3D Gaussians are fully learnable and can be directly optimized through end-to-end training via view reconstruction.

\subsection{Data Preprocessing}    
\label{subsec:data_preprocess}
Our system utilizes LiDAR scans and RGB images as inputs, incorporating a robust preprocessing pipeline to enhance data quality and accuracy. As shown in the data processing part in Fig.~\ref{fig:big_diagram}, initially, we employ Fast-livo2 \cite{zheng2024fast}, a state-of-the-art, deeply coupled LiDAR-inertial-visual SLAM system, to estimate initial LiDAR scan poses and generate a dense point cloud map. 

Next, the point cloud undergoes denoising, downsampling, and normal estimation. It is then converted into a mesh using Poisson Surface Reconstruction \cite{kazhdan2006poisson}. A cleaned point cloud is further filtered based on its spatial distance to the reconstructed mesh. At the same time, depth images $D_L$ and normal images $N_L$ are also generated from point cloud map through 3D-to-2D projection.

\subsection{Initialization}
\label{subsec:initialization}

Building upon previous work \cite{lu2024scaffold}, we propose an anchor-based and normal-assisted Gaussian initialization method for colorized point cloud generated in Sec.~\ref{subsec:data_preprocess}. Unlike traditional methods such as 3DGS vanilla \cite{kerbl3Dgaussians} and Scaffold-GS \cite{lu2024scaffold}, which initialize Gaussians with identity quaternions and uniform scaling, our approach utilizes normal vectors from dense SLAM-generated point clouds to rotate and shape Gaussian ellipsoids. To the best of our knowledge, this is the first method to apply anchor-based and normal-assisted Gaussian initialization for constructing 3DGS maps. 



\subsubsection{Voxelization and Anchor-based Point Generation} We first voxelize the scene from the input point cloud $\mathbf{P} \in \mathbb{R}^{M \times 3}$. Specifically, the raw points are discretized into a regular 3D voxel grid, from which anchor-level point features are generated.
\begin{equation}
    \mathbf{P'}= \{ p_i \in P \mid \lfloor p_i / \epsilon \rfloor \text{ is unique} \}
\end{equation}
where \( \epsilon \) is the voxel size, and \( \mathbf{P'} \in \mathbb{R}^{N \times 3} \) represents sampled points by the voxel size.  Each point \( p \in \mathbf{P'} \) is treated as an anchor pointfor generating a Gaussian ellipsoid, equipped with a local context feature $f_v \in \mathbb{R}^{32}$, a scaling factor $l_v \in \mathbb{R}^3$, and $k$ learnable offsets $O_v \in \mathbb{R}^{k \times 3}$.

\subsubsection{Normal-Assisted Gaussian Initialization}
To achieve precise alignment with surface geometry, we flatten Gaussian ellipsoids along the normal direction by rotating it to align with the surface normal and squeezing the Gaussian ellipsoid scale along a specific direction. The alignment of a Gaussian's principal axis with a normal vector is achieved through quaternion-based rotation.

\paragraph{Gaussian Ellipsoid Rotation}
Given an initial Gaussian principal axis $\mathbf{v}_1 = (0,0,1)$, the quaternion $q$ is computed as:
\begin{equation}
q \cdot \mathbf{v}_1 \cdot q^{-1} = \mathbf{n}
\end{equation}
where $\mathbf{n} = (n_x, n_y, n_z) \in \mathbb{R}^3$ is the target normal vector from the point cloud in world coordinates.

The axis of rotation is derived from the cross-product:
\begin{equation}
\mathbf{a} = \mathbf{v}_1 \times \mathbf{n}
\end{equation}

which produces a vector perpendicular to both directions. 
To ensure numerical stability, the axis is normalized as $\mathbf{a}_{\text{norm}} = (a_x, a_y, a_z)$.
    
The rotation angle $\theta$ between $\mathbf{v}_1$ and $\mathbf{n}$ is calculated as:
\begin{equation}
\theta = \cos^{-1} (\mathbf{v}_1 \cdot \mathbf{n})
\end{equation}

Using this angle, the rotated quaternion is constructed as:
\begin{equation}
q = \left( \cos \frac{\theta}{2}, \sin \frac{\theta}{2} \cdot a_x, \sin \frac{\theta}{2} \cdot a_y, \sin \frac{\theta}{2} \cdot a_z \right)
\end{equation}

\paragraph{Gaussian Ellipsoid Flattening}
After aligning the Gaussian with the normal vector, we apply flattening along the normal direction. The original Gaussian ellipsoid scales are:
\[
\mathbf{s} = \log \left( \sqrt{\|\mathbf{d}^2\|} \right)  , \quad \mathbf{s} \in \mathbb{R}^3
\]
To enforce flattening, only $s_3$ in $\mathbf{s} = (s_1,s_2,s_3)$ is updated using the flattening factor $\sigma_f$:
\begin{equation}
s_3 = \sigma_f * s_3
\end{equation}

\begin{figure}[h]
\centering
  \begin{subfigure}[b]{0.23\textwidth}
    \includegraphics[trim={0cm 0cm 0cm 0cm},clip, width=\textwidth]{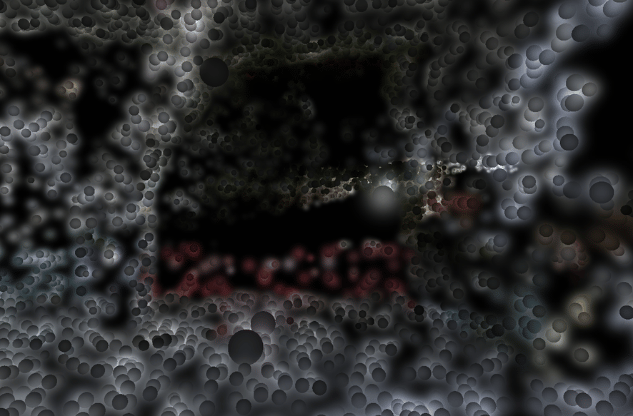}
  \end{subfigure}
  \begin{subfigure}[b]{0.23\textwidth}
    \includegraphics[trim={0cm 0cm 0cm 0cm},clip, width=\textwidth]{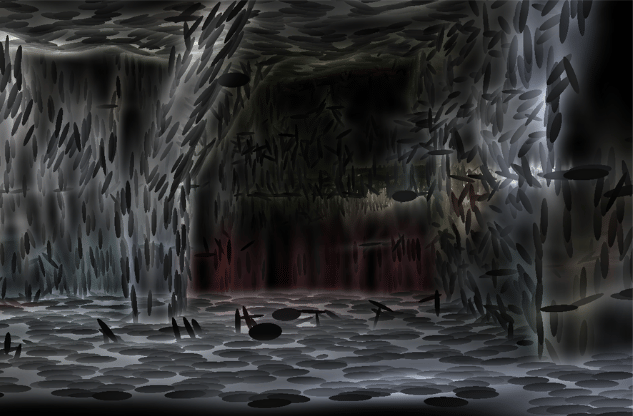}
  \end{subfigure}
\caption{Sample of Gaussian initialization using vanilla 3DGS initialization (left) and normal-assisted Gaussian initialization (right)}
\label{fig:gaussian_initialize}
\end{figure}

The effect of our initialization strategy is illustrated in Fig.~\ref{fig:gaussian_initialize}, where Gaussian ellipsoids are flattened and aligned with the surface.

\subsection{Optimization}
\label{subsec:optimization}
Upon LiDAR-enhanced initialization, we apply a comprehensive regularization strategy comprising Photometric Loss, Flatten Loss, Offset Regularization Loss, Depth Loss, and Normal Loss.

\subsubsection{Photometric loss ($\mathcal{L}_{\text{RGB}}$)}
Same with vanilla 3DGS \cite{kerbl3Dgaussians}, we minimize the difference between the rendered RGB image $\Tilde{I}$ and the input image $I_{gt}$ via
 \begin{equation}
     \mathcal{L}_{RGB} = (1 - \lambda) L_1 (\Tilde{I},I_{gt}) + \lambda L_{D-SSIM}(\Tilde{I},I_{gt})
 \end{equation}
where $\lambda$ is D-SSIM weight, we use $\lambda = 0.2$ in all our experiments..
\subsubsection{Flatten loss ($\mathcal{L}_{\text{f}}$)}
Traditional 3DGS methods often position Gaussian centers inside the surface, limiting surface regularization. Our approach refines 3D Gaussian ellipsoids into highly flattened shapes, reducing overlap and aligning them closely with the surface. The scaling factor $\mathbf{s}=(s_1,s_2,s_3)$ defines the ellipsoid's dimensions. The flattening loss is defined as:
\begin{equation}
    \mathcal{L}_{f} = \|\mathbf{s_3}\|_2
\end{equation}

where $\mathbf{s_3}$ indicates scale of Gaussian scale along normal direction.

\subsubsection{Offset loss ($ \mathcal{L}_{\text{o}}$)}
Given the confidence in the accuracy of the input LiDAR point cloud, we introduce constraints on the learnable offset factors. The offset factor $\mathbf{O}=(o_1,o_2,o_3)$ defines the ellipsoid's offset dimensions. To constrain the offset of Gaussian ellipsoids along the surface, we introduce an offset product loss on $o_1$ and $o_2$. Additionally, to minimize the offset of Gaussian ellipsoids along the normal direction, we also minimize the z-axis component $o_3$ in $\mathbf{O}$: 
\begin{equation}
    \mathcal{L}_o = \lambda_{o_3} \|\mathbf{o}_3\|_1 + \lambda_{o_{1,2}} \|\mathbf{o}_1 \cdot \mathbf{o}_2 \|_1
\end{equation}
where $ \lambda_{o_3}$ and $\lambda_{o_{1,2}}$ are weights for the offset along normal direction and against normal direction. Since we aim to impose stronger spatial constraints along the normal direction, a higher weight is assigned to $\lambda_{o_3} = 5$, while $\lambda_{o_{1,2}} = 1$.

\subsubsection{Depth loss}
\cite{hierarchicalgaussians24}
We adopt the depth loss approach from Sparse-GS \cite{xiong2023sparsegs}, comparing the LiDAR depth image $D_L$ (Sec.~\ref{subsec:data_preprocess}) with the rendered depth image  $\Tilde{D}$. To filter out empty and invalid depth values from LiDAR, the depth mask ${M}_D$ is calculated as:
\begin{equation}
{M}_D(i, j) = \begin{cases} 
1 & \text{if } D_L(i, j) > T_{\text{min}} \And D_L(i, j) < T_{\text{max}}  \\ 
0 & \text{otherwise} 
\end{cases}
\end{equation}
The depth loss is then:
 \begin{equation}
     \mathcal{L}_{D}= L_1 (inv(\Tilde{D}),inv(D_{L})) * {M}_D
 \end{equation}

\subsubsection{Normal loss ($\mathcal{L}_{\text{N}}$)}


To further enforce geometric consistency, we introduce normal consistency regulariziation loss. This consistency normal loss leverages the curvature map \(|\nabla\boldsymbol{\Tilde{N}}|\in [0,1]^{H \times W}\), derived from the rendered normal image $\Tilde{N}$, to promote smoothness while preserving high-frequency details using an edge image $E$ from the 3D LiDAR point cloud by \cite{lu2019fast}. Similar to depth loss, a normal mask $M_N$ is also used to filter out pixels without normal values.
\begin{equation}
    \mathcal{L}_{N} =  (\nabla\boldsymbol{\Tilde{N}} \odot \mathbf{w}(E)) * M_N 
\end{equation}
where the weight function $\mathbf{w}(x) = {(x - 1)}^{q}$ is designed to weight the gradient in terms of the edges in $E$. We use $q = 400$ in our experiments.

\subsubsection{Composite Loss Function ($\mathcal{L}_{G}$)}
Finally, the composite loss function is formulated as:
\begin{equation}
    \mathcal{L}_{G} = \lambda_{c} \mathcal{L}_{\text{RGB}} + \lambda_f \mathcal{L}_f + \lambda_o \mathcal{L}_{\text{o}} + \lambda_D \mathcal{L}_{\text{D}} + \lambda_{N}\mathcal{L}_{N}
\end{equation}
where $\lambda_c$, $\lambda_f$, $\lambda_o$, $\lambda_D$ and $\lambda_N$ control the weighting of flatten, offset, depth, and normal loss terms, respectively. With this mutual optimization, the point cloud generated from 3D Gaussians will be refined, which in turn provides more reliable prior.

\section{Experiment Dataset}
This section introduces datasets for validating Structured-Li-GS: method. The dataset covers diverse scenarios, including sparse-view, outdoor, and unbounded scenes.

\subsection{Public Dataset}
We use the FASTLIVO2 benchmark dataset \cite{zheng2024fast}, a robust and open-source dataset designed for LiDAR–visual SLAM research. It provides 1280×1024 RGB image sequences captured with an MV-CA013-21UC industrial camera, accompanied by LiDAR scans and IMU measurements from a Livox Avia sensor, which offers a wide field of view and high point density. We evaluate five benchmark sequences: CBD Building 2, Red Sculpture, Retail Street, SYSU, and Main Building. All experiments are conducted using point cloud maps and posed images generated by the LiDAR–visual SLAM pipeline. Configurations of the sequences can be found in Table.~\ref{tab:dataset}.

We further evaluate on the Hilti’22 dataset \cite{9968057}, which contains indoor and outdoor sequences collected using both handheld and robot-mounted platforms in diverse environments such as construction sites, offices, laboratories, and parking areas. These sequences present numerous challenges, including long corridors, basements, staircases, textureless surfaces, varying illumination, and sparse LiDAR plane constraints. The handheld platform uses a Hesai PandarXT-32 LiDAR at 10 Hz, five wide-angle cameras at 40 Hz (downsampled to 10 Hz), and a Bosch BMI085 IMU at 400 Hz. The robot-mounted platform employs a Robosense BPearl5 LiDAR at 10 Hz, eight omnidirectional cameras at 10 Hz, and an Xsens MTi-670 IMU at 200 Hz. For both platforms, only the front-facing camera is used for the systems evaluated in this work. We evaluate three benchmark sequences, including Exp04 (Construction Upper Level), Exp14 (Basement 2), and Exp21 (Outside Building). Configurations of these sequences can be found in Table.~\ref{tab:dataset}.

\begin{figure}[] 
    \centering
    \includegraphics[width=0.23\textwidth]{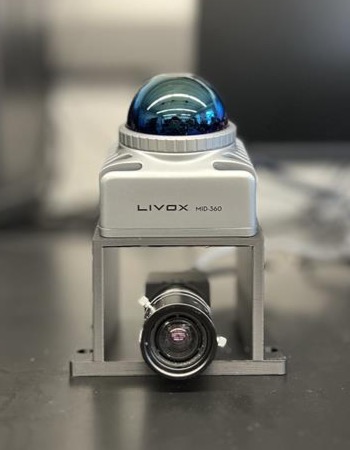} 
    \includegraphics[width=0.23\textwidth]{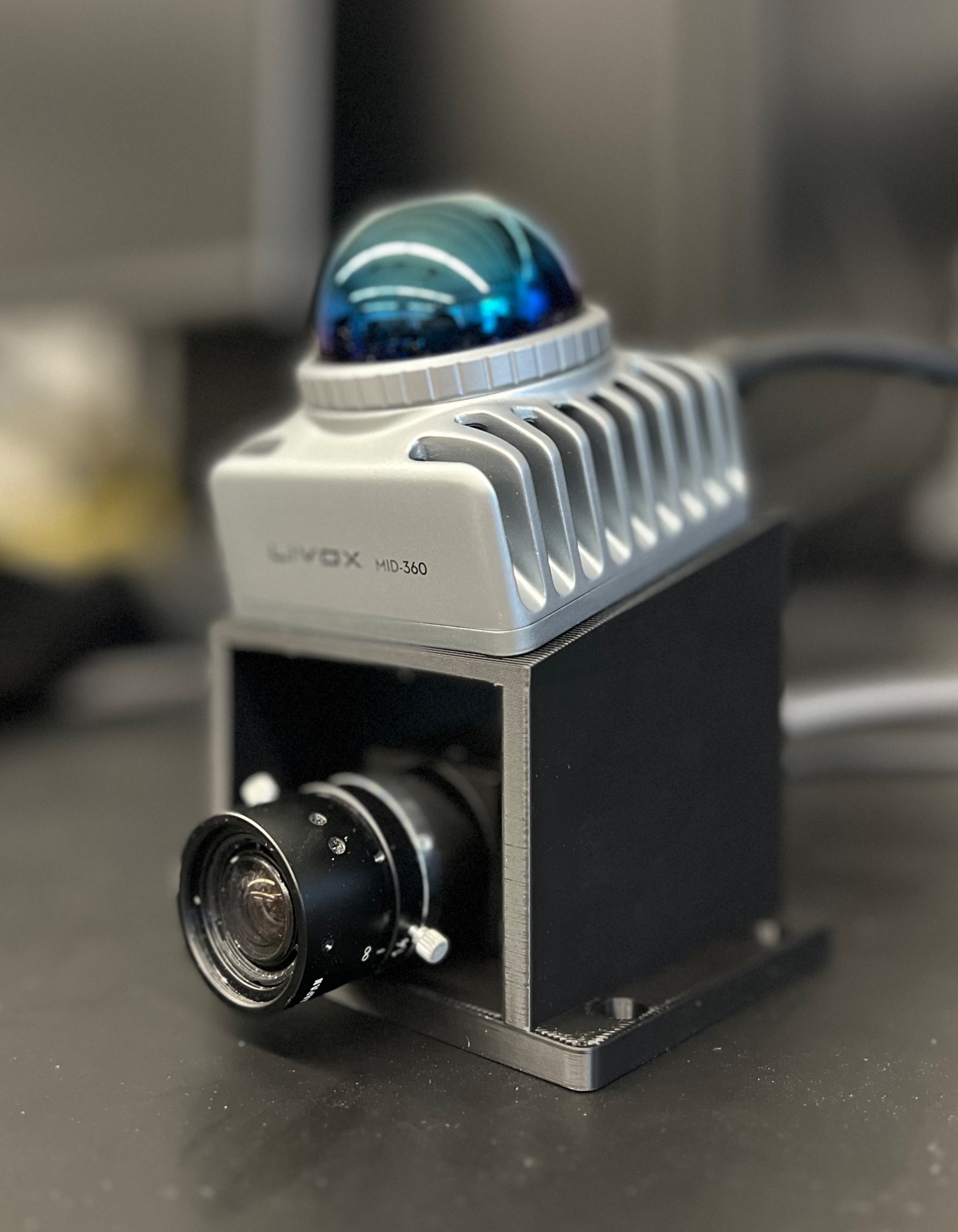}
    \caption{Custom handheld LiDAR-Camera scanner with hardware synchronization for data acquisition}
    \label{fig:scanner}
\end{figure}

\begin{table}[t]
\centering
\setlength{\tabcolsep}{4pt}
\small
\caption{Summary of datasets and their sequences used for experimental evaluation, including the number of images, initial point cloud size, and approximate trajectory distance for each sequence. This setup spans three categories: FASTLIVO2, the HILTI2022 dataset, and our custom-captured sequences.}
\label{tab:dataset}
\begin{tabular}{llccc} 
\toprule
Dataset & Seq & \#Img & Init.\#Pts  \\
\midrule
\multirow{5}{*}{FASTLIVO2}
  & CBD2          & 444   & 356,885   \\
  & Retail Street & 450   & 245,983    \\
  & Red Sculpture &  339  & 1,621,509   \\
  & SYSU & 246   & 479,776   \\
  & Main Building & 209 & 362,958\\
\midrule
\multirow{3}{*}{HILTI22}
  & Exp04 Construction & 284 & 822,483\\
  & Exp14 Basement & 103 & 90,070   \\
  & Exp21 Outside Building & 466 & 438,600\\
\midrule
\multirow{2}{*}{Custom}
  & Corride & 424   & 1,325,242   \\
  & EIT  & 307 & 415,924  \\
\bottomrule
\end{tabular}
\end{table}

\begin{table*}[]
\centering
\small
\setlength{\tabcolsep}{3pt}
\caption{Rendering quality comparison across different sequences of the Fast-LIVO2 dataset. We report PSNR ($\uparrow$), LPIPS ($\downarrow$), and SSIM ($\uparrow$) on the split test set. The best score for each metric is shown in \textbf{bold}, and the second-best is \underline{underlined}.}
\label{tab:comparison_table}
\begin{tabular}{lccc|ccc|ccc|ccc|ccc}
\toprule
\multicolumn{1}{c}{\text{Dataset}} & \multicolumn{15}{c}{FASTLIVO2 Dataset} \\ \cmidrule(lr){2-16}
\multicolumn{1}{c}{\text{Squence}} & \multicolumn{3}{c|}{\text{CBD2}} & \multicolumn{3}{c|}{\text{Red Sculpture}} & \multicolumn{3}{c|}{\text{Retail Street}}& \multicolumn{3}{c|}{\text{SYSU}} & \multicolumn{3}{c}{\text{Main Building}}  \\

\text{Method} & \text{PSNR}  & \text{SSIM}  & \text{LPIPS}   & \text{PSNR}  & \text{SSIM}  & \text{LPIPS}    & \text{PSNR} & \text{SSIM}  & \text{LPIPS}  & \text{PSNR}  & \text{SSIM}  & \text{LPIPS} & \text{PSNR}  & \text{SSIM}  & \text{LPIPS}   \\
\midrule
    \text{3D-GS} & 19.11 &  0.703 & 0.444 & 23.83 & 0.743 & 0.271 & \textbf{28.87} & \textbf{0.907} & \textbf{0.099} & 23.65 & 0.718 & 0.402 & 22.83 & 0.700 & 0.348 \\
    \text{2D-GS}  & {20.82} & 0.723 &  0.447 & 23.53 & 0.739 & 0.301 & 27.81 & 0.891 & 0.124 & 23.25 & 0.709 & 0.426  & 22.90 & 0.714 & 0.365 \\
    \text{Scaffold-GS} & \underline{22.75} & \underline{0.760} & \underline{0.390} & {24.88} & \textbf{0.781} &  \underline{0.234} & 28.66 & {0.897} &  \underline{0.100} & 24.26 & \textbf{0.733} & 0.374  & \underline{25.94} & \underline{0.755} & \underline{0.289}\\
    \text{LetsGo}  & 22.57 & \textbf{0.788} & 0.425 & 21.80 & 0.712 & 0.331 & 27.64 & 0.895 & 0.120 & 23.03 & 0.701 & 0.419  & 21.90 & 0.702 & 0.374\\
    \text{AtomGS}  & 21.57 & 0.739 & 0.402 &  \textbf{23.94} & 0.743 & 0.288 & 28.43 & 0.890 & 0.125 & \underline{24.35} & \underline{0.729} & \underline{0.373} & 22.24 & 0.686 & 0.375\\
    \midrule
    \text{Ours}  & \textbf{22.82} & \underline{0.760} & \textbf{0.389} &  \underline{24.89} & \underline{0.761} & \textbf{0.223} &  \underline{28.83} & \underline{0.899} & 0.117 & \textbf{24.79} & \textbf{0.733} & \textbf{0.362}  & \textbf{26.15} & \textbf{0.763} & \textbf{0.288}\\
\bottomrule
\end{tabular}
\end{table*}

\begin{table*}[]
\centering
\small
\setlength{\tabcolsep}{2.8pt}
\caption{Rendering Quality Comparison on different sequences from HILTI22 and custom dataset.  We report PSNR ($\uparrow$), LPIPS ($\downarrow$), and SSIM ($\uparrow$)on the split test set. The best performance for each metric is highlighted in \textbf{bold}, and the second-best is \underline{underlined}.}
\label{tab:comp_hilti_custom}
\begin{tabular}{lccc|ccc|ccc|ccc|ccc}
\toprule
\multicolumn{1}{c}{\text{Dataset}} & \multicolumn{9}{c}{HILTI22}  & \multicolumn{6}{c}{Custom Dataset} \\ 
\cmidrule(lr){2-10} \cmidrule(lr){11-16} 
\multicolumn{1}{c}{\text{Squence}}  & \multicolumn{3}{c|}{\text{Exp04 Construction}}  & \multicolumn{3}{c|}{\text{Exp14 Basement}} & \multicolumn{3}{c|}{\text{Exp21 Outside Building}} & \multicolumn{3}{c|}{\text{Corride}}& \multicolumn{3}{c}{\text{EIT Building}} \\

\text{Method} & \text{PSNR}  & \text{SSIM}  & \text{LPIPS}   & \text{PSNR}  & \text{SSIM}  & \text{LPIPS}    & \text{PSNR} & \text{SSIM}  & \text{LPIPS}  & \text{PSNR}  & \text{SSIM}  & \text{LPIPS} & \text{PSNR}  & \text{SSIM}  & \text{LPIPS}  \\
\midrule
    \text{3D-GS} & 22.00 & 0.792 & 0.452 & \textbf{38.65} & \textbf{0.945} & \textbf{0.164} & 23.94 & 0.837 & 0.321 & 24.41 & 0.764 & \underline{0.470} & 22.18 & 0.752 & \underline{0.479} \\
    \text{2D-GS} & 22.02 & 0.796 & 0.475 & 38.00 & \underline{0.943} & 0.172 & 23.34 & 0.820 & 0.358 & 23.63 & 0.755 & 0.516 & 21.63 & 0.747 & 0.512\\
    \text{Scaffold-GS} & \underline{23.67} & \textbf{0.841} & \underline{0.389} & 37.64 & 0.923 & 0.191 & \underline{24.16} & \textbf{0.833} & \underline{0.317} & \underline{25.41} & \textbf{0.779} &  \underline{0.470} & \underline{23.32} & \textbf{0.772} & 0.485\\
    \text{AtomGS} & 23.20 & 0.810 & 0.448 & 38.04 & 0.942 & 0.178 & 23.99 & 0.826 & 0.343 & 25.16 & 0.768 & 0.481 & 22.76 & 0.754 & 0.480\\
    \midrule
    \text{Ours} & \textbf{23.96} & \underline{0.839} & \textbf{0.396} & \underline{38.43} & 0.940 &  \underline{0.169} &  \textbf{24.31} & \underline{0.832} & \textbf{0.312} & \textbf{25.58} & \underline{0.773} & \textbf{0.457} & \textbf{23.97} & \underline{0.770} & \textbf{0.478}\\
\bottomrule
\end{tabular}
\end{table*}

\subsection{Custom Dataset}
To evaluate the system’s performance under real-world conditions (e.g., low texture, long-range scenes), we collect a new custom dataset. Our data collection platform, illustrated in Fig.~\ref{fig:scanner}, is equipped with an industrial camera (MV-CA013-21UC), a Livox MID360 LiDAR, and a NUC 11 mini-PC (Intel i7-1165G7 CPU and 8 GB RAM) as the onboard computer. The camera provides a FoV of $70.6^\circ \times 68.5^\circ$, while the LiDAR offers a FoV of $360^\circ \times 59^\circ$. All sensors are hardware-synchronized using Teensy-based timers: the camera is triggered at 10 Hz, and the LiDAR is triggered at 1 Hz along with GPRMC messages. The private dataset contains two sequences collected in structure-less and weakly textured environments, Corride and EIT building. Configuration of the sequences are shown in Table.~\ref{tab:dataset}.

\begin{figure}
    \centering
    \includegraphics[width=\linewidth]{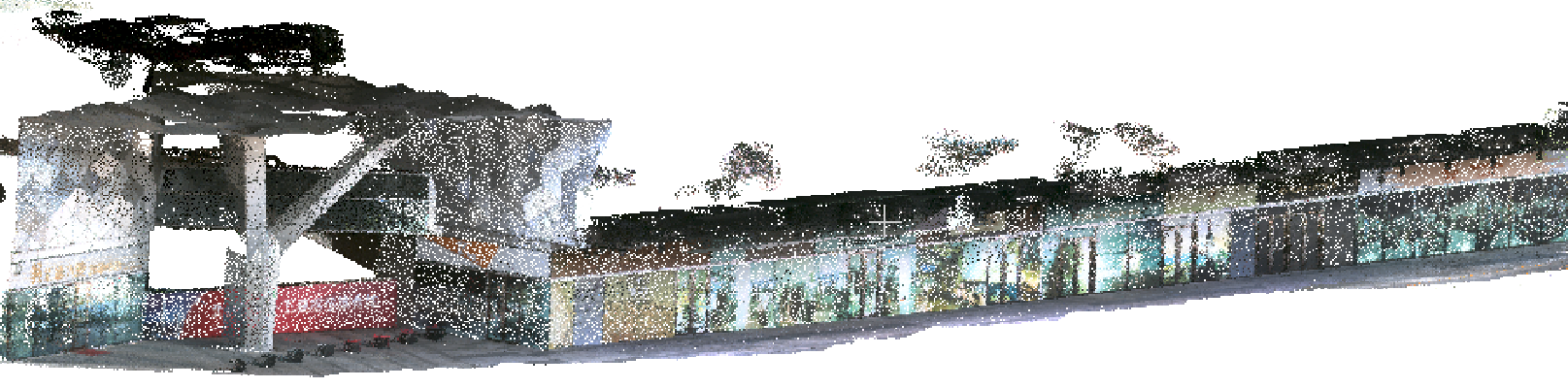}
    \includegraphics[width=\linewidth]{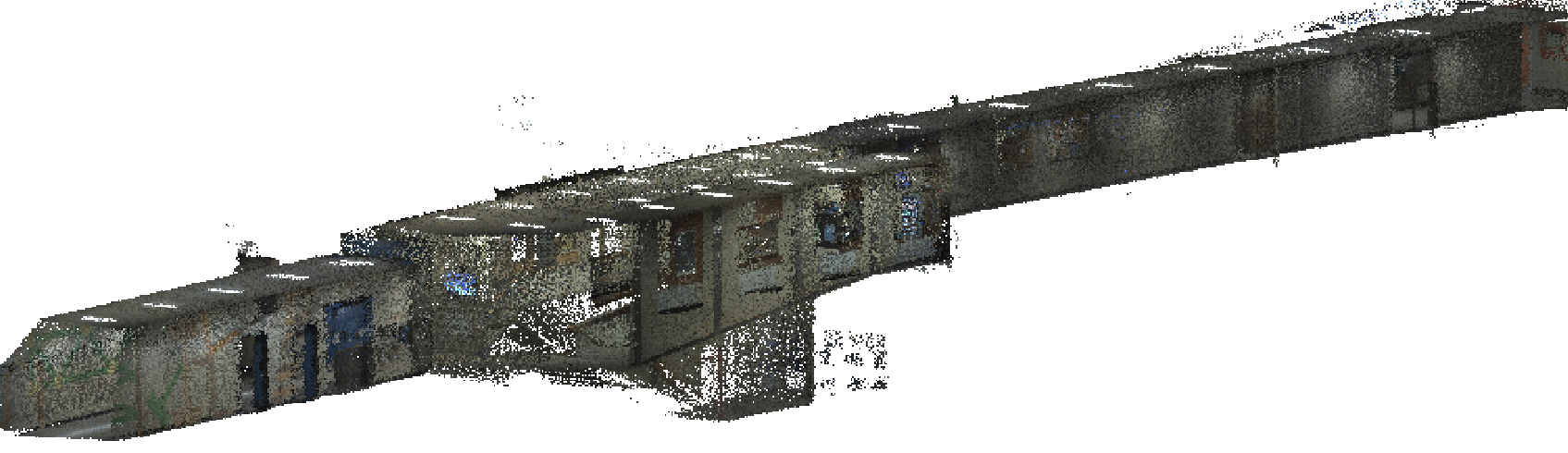}
    \caption{Examples of colorized 3D point clouds generated during SLAM-based preprocessing: CBD2 sequence from FASTLIVO2 dataset (Above), and Corride sequence from Custom dataset (Bottom).}
    \label{fig:pts_display}
\end{figure}

\begin{figure*}[h]
    \centering
    \renewcommand{\arraystretch}{0.8} 
    \setlength{\tabcolsep}{1pt} 
    \captionsetup{belowskip=0pt} 
    \begin{tabular}{ccccc}
         \textbf{Ground Truth} &\textbf{Structured-Li-GS (Ours)} & \textbf{Scaffold-gs} & \textbf{LetsGo}& \textbf{3DGS} \\
        \includegraphics[width=0.18\linewidth]{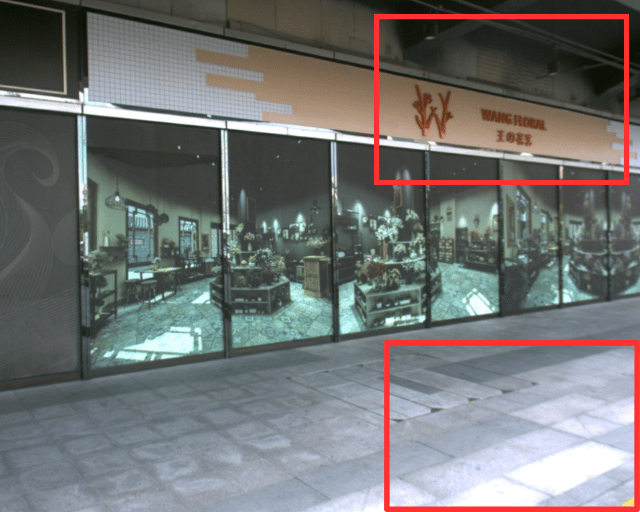} &
        \includegraphics[width=0.18\linewidth]{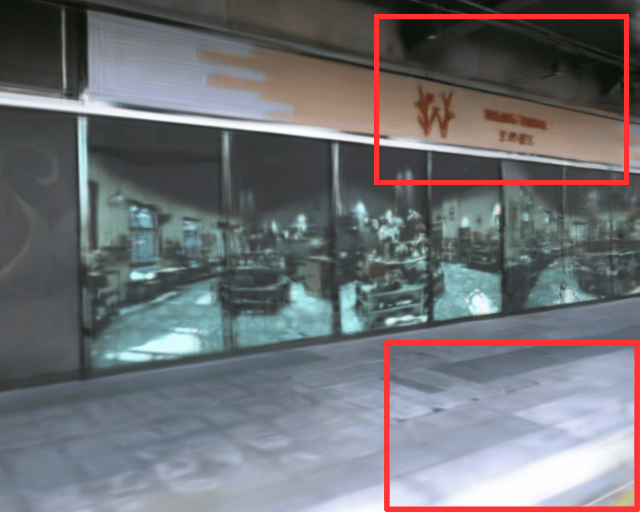}&
        \includegraphics[width=0.18\linewidth]{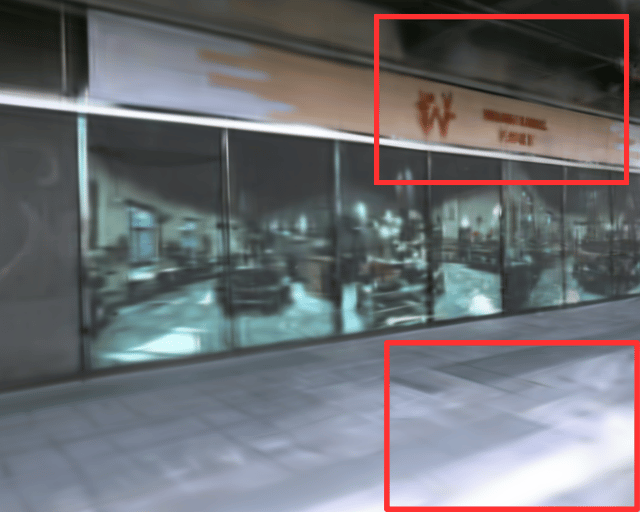} &
        \includegraphics[width=0.18\linewidth]{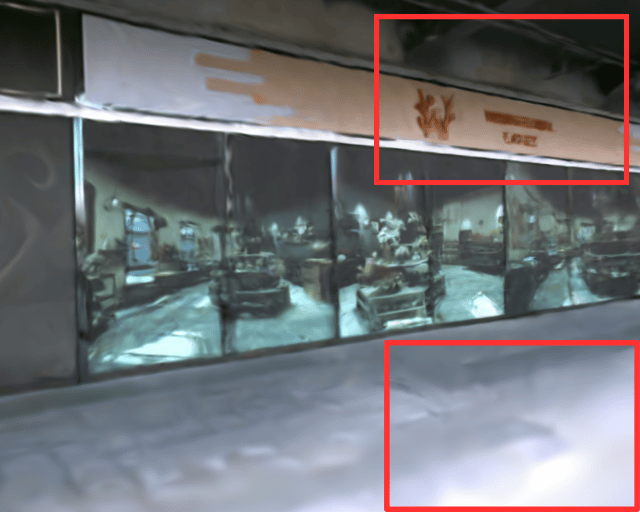}&
        \includegraphics[width=0.18\linewidth]{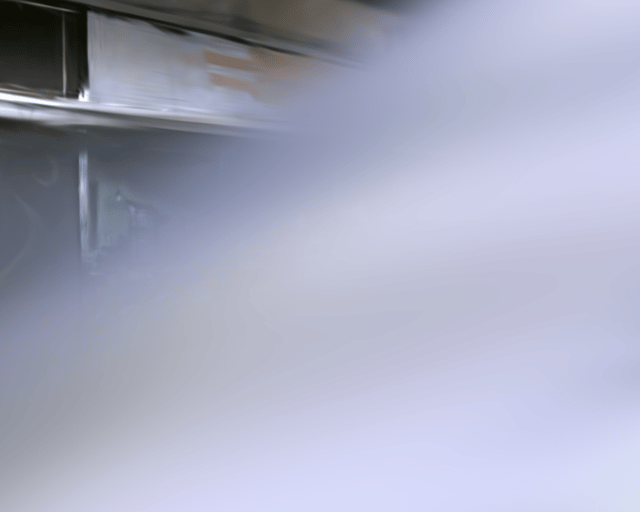}\\
        \includegraphics[width=0.18\linewidth]{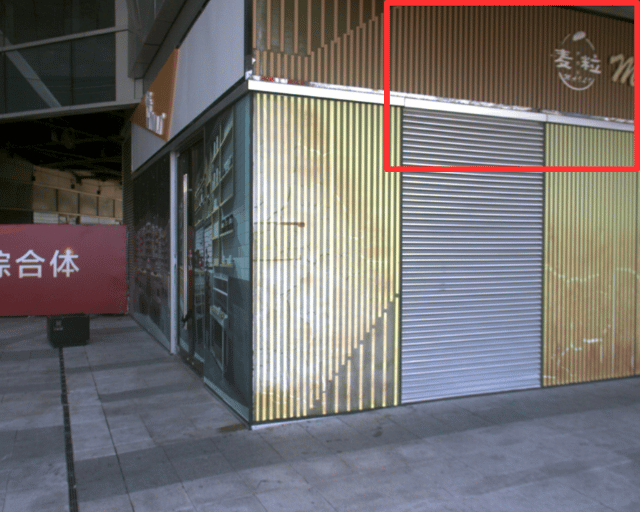} &
        \includegraphics[width=0.18\linewidth]{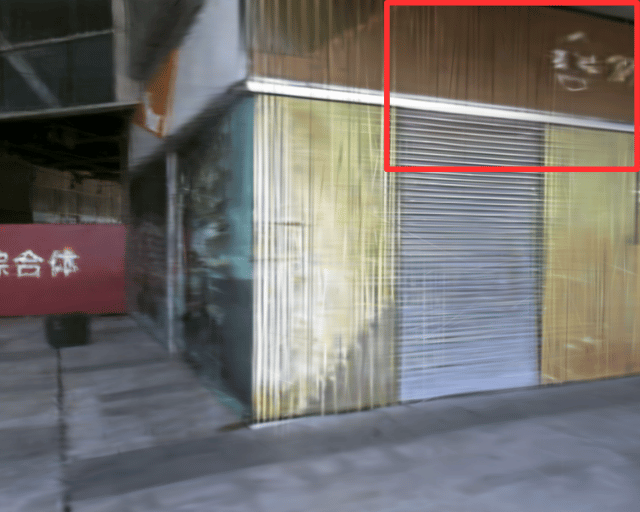} &
        \includegraphics[width=0.18\linewidth]{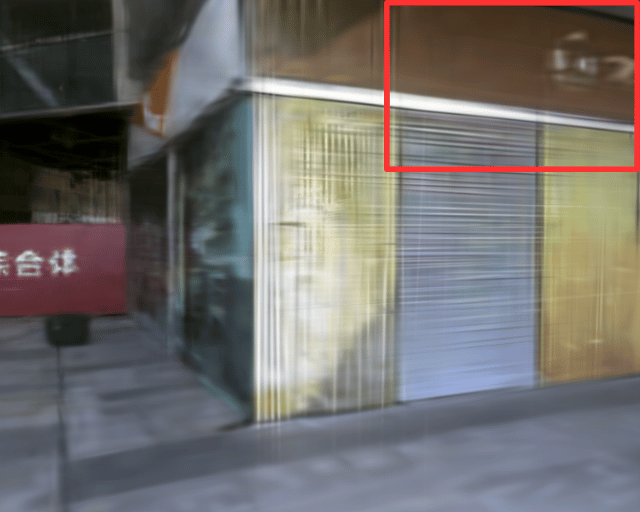} &
        \includegraphics[width=0.18\linewidth]{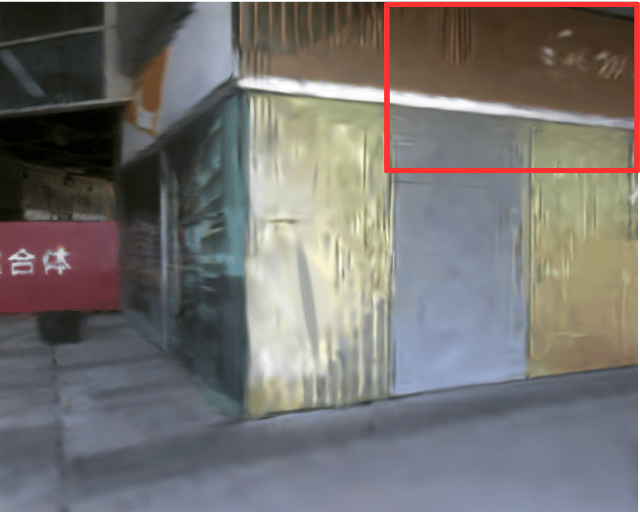}&
        \includegraphics[width=0.18\linewidth]{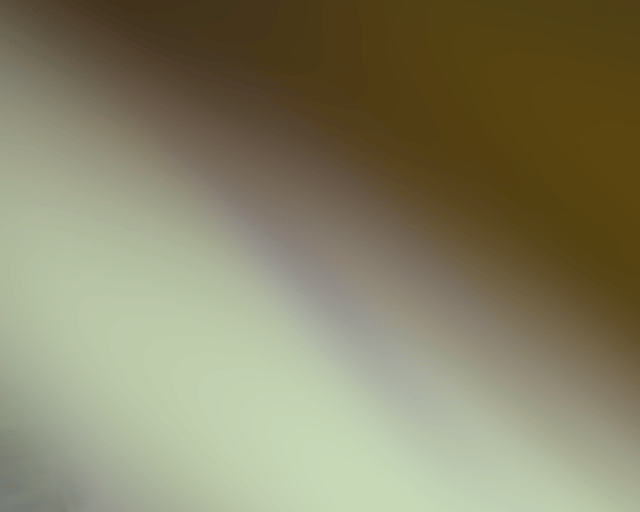}\\
        \includegraphics[width=0.18\linewidth]{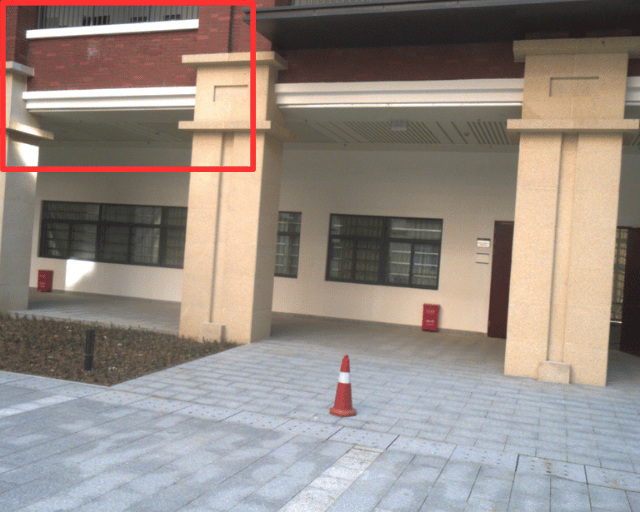} &
        \includegraphics[width=0.18\linewidth]{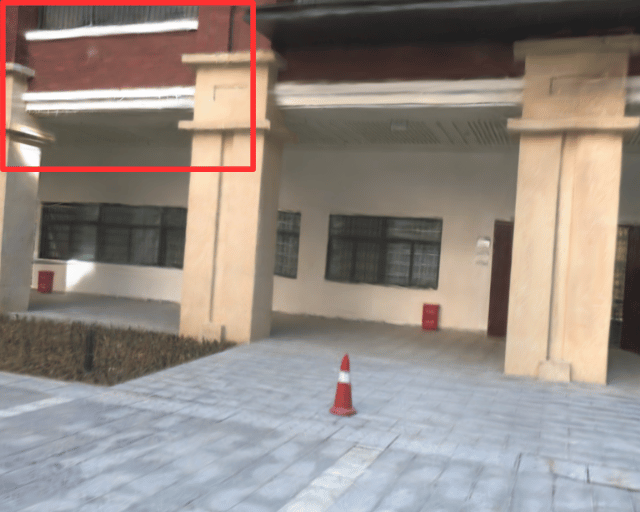} &
        \includegraphics[width=0.18\linewidth]{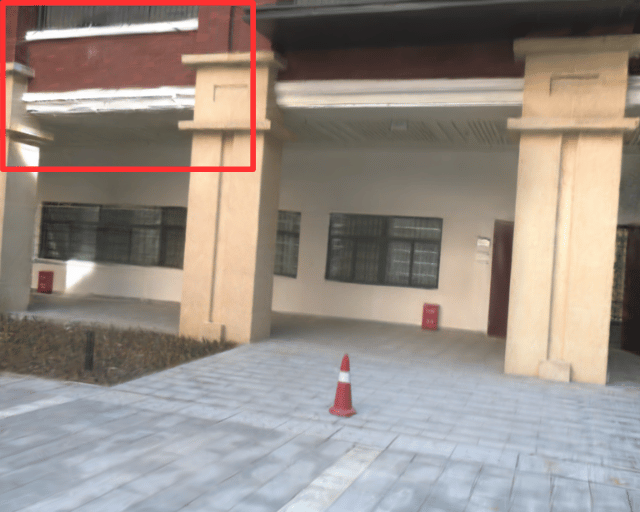} &
        \includegraphics[width=0.18\linewidth]{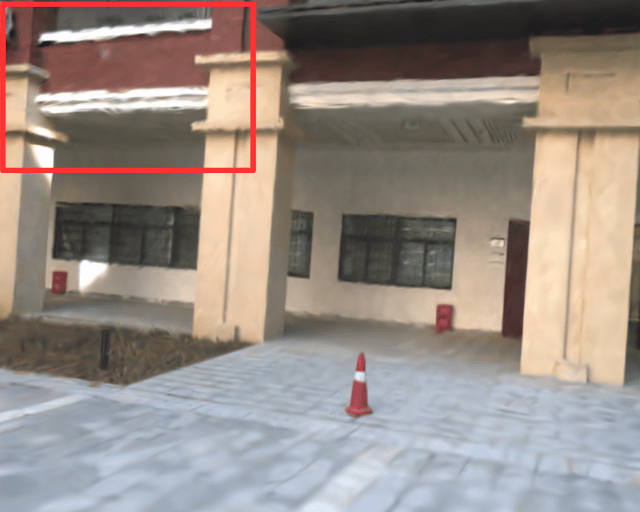}&
        \includegraphics[width=0.18\linewidth]{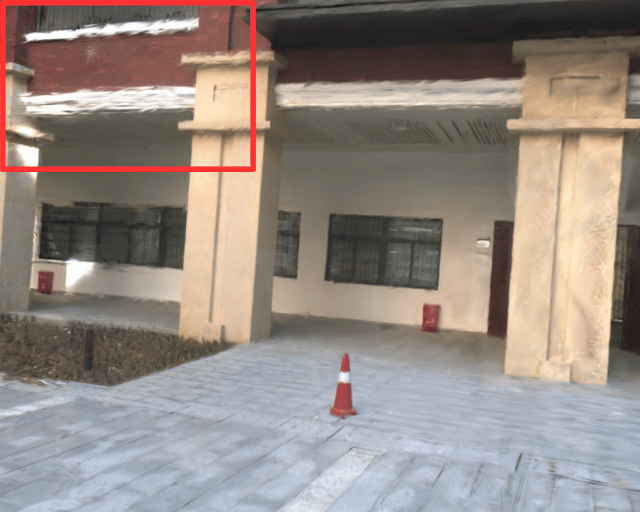}\\
        \includegraphics[width=0.18\linewidth]{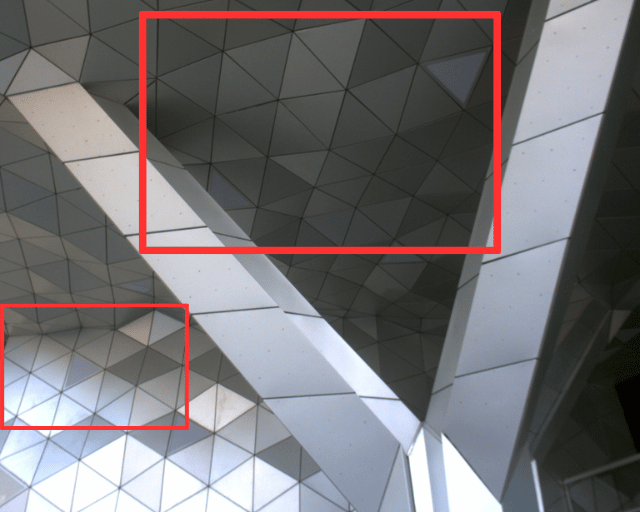} &
        \includegraphics[width=0.18\linewidth]{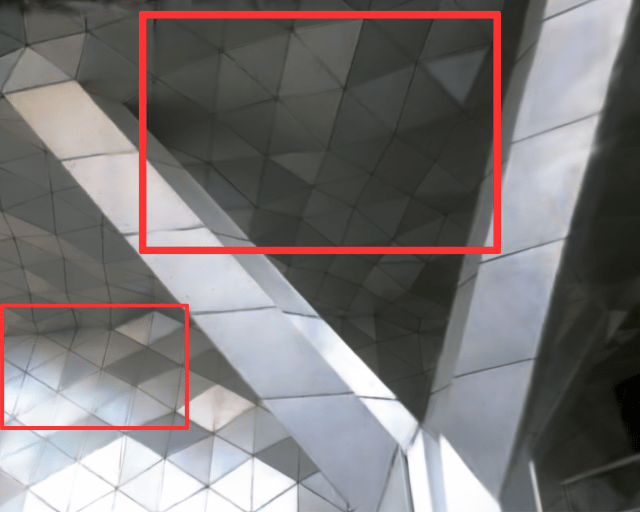} &
        \includegraphics[width=0.18\linewidth]{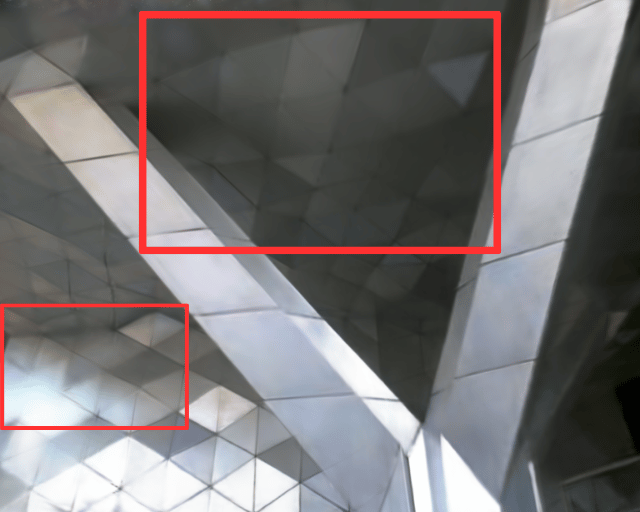} &
        \includegraphics[width=0.18\linewidth]{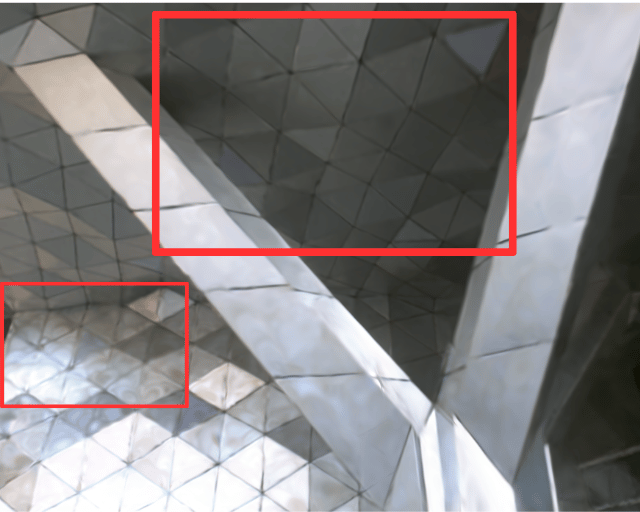}&
        \includegraphics[width=0.18\linewidth]{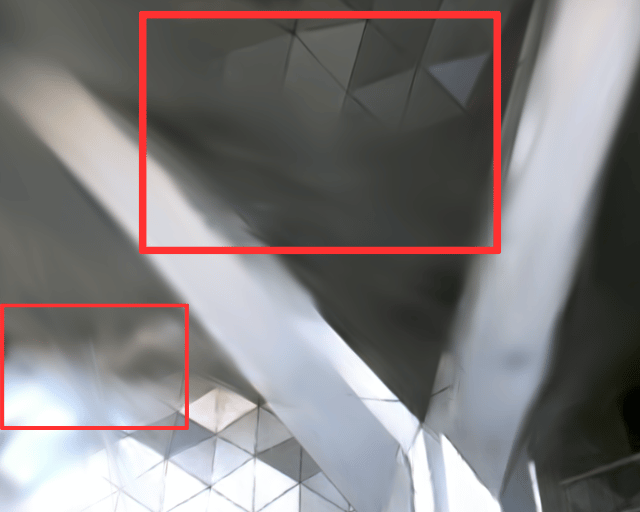}\\
    \end{tabular}
    \caption{Qualitative comparison of rendered images from different Gaussian Splatting methods on FAST-LIVO2 dataset. Red boxes highlight key regions where our method preserves fine geometric and photometric details more accurately than competing models.}
    \label{fig:comparison_figures}
\end{figure*}

\section{Experiment}
This section evaluates Structured-Li-GS: method, in terms of its rendering quality against previous state-of-the-art approaches (SOTA). Before rendering evaluation, benchmark datasets and custom datasets are dealed with FASTLIVO2 algorithm and following operations in pre-processing stage. Colorized 3D point clouds, as shown in Fig.~\ref{fig:pts_display}, and poses images (RGB, depth images) are prepared for Gaussian Splatting training.

Rendering quality was evaluated using standard image comparison metrics:(1) Peak Signal-to-Noise Ratio (PSNR), (2) Structural Similarity Index Measure (SSIM) \cite{wang2004image}, (3) Learned Perceptual Image Patch Similarity (LPIPS). We benchmark our approach against several methods image-based 3DGS , (vanilla 3DGS\cite{kerbl3Dgaussians}, AtomGS\cite{liu2024atomgs}, Scaffold-GS \cite{lu2024scaffold}), planar Gaussian (2DGS\cite{Huang2DGS2024}), as well as SOTA LiDAR-visual fusion 3DGS reconstruction methods (LetsGo\cite{letsgo}).

\subsection{Experimental Setup}
To enhance comparison efficiency, we subsampled the input point cloud and downsampled the inputed images by a different factor for different datasets. For FASTLIVO2 and HILTI22 dataset, we subsampled the input point cloud for training GS models using a voxel size of 0.06m, and subsampled the images by a factor of 5. For Private dataset, we subsampled the input point cloud for training GS models using a voxel size of 0.06m, and subsampled the images by a factor of 3. 


Furthermore, we employ a train/test split by reserving every 8\textsuperscript{th} image within each dataset sequence for testing. All experiments utilize identical loss function weights, set as $\lambda_c = \lambda_o = \lambda_D = \lambda_{N} = 1$, to ensure balanced optimization across all objectives without biasing the training toward any single term. This uniform weighting facilitates fair comparisons between different datasets and configurations by maintaining consistent learning dynamics across experiments. All experiments were conducted on a workstation equipped with an NVIDIA RTX A5000 GPU and an AMD Ryzen Threadripper PRO 5955WX 16-core processor.

\begin{figure*}[h]
    \centering
    \renewcommand{\arraystretch}{0.8} 
    \setlength{\tabcolsep}{1pt} 
    \captionsetup{belowskip= 0pt} 
    \begin{tabular}{ccccc}
         \textbf{Ground Truth} &\textbf{Structured-Li-GS (Ours)} & \textbf{Scaffold-gs}& \textbf{2DGS} & \textbf{3DGS} \\
        \includegraphics[width=0.18\linewidth]{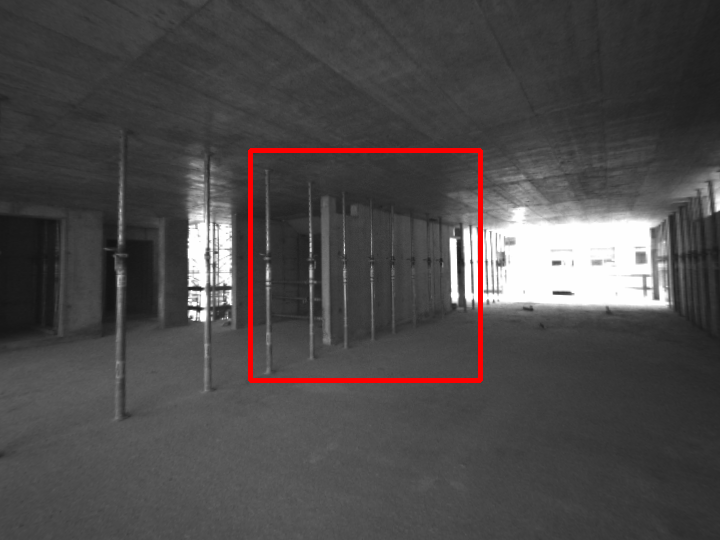} &
        \includegraphics[width=0.18\linewidth]{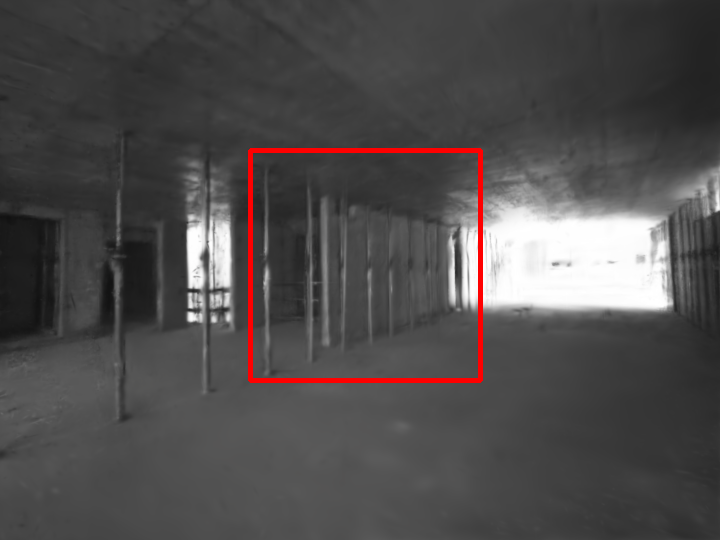}&
        \includegraphics[width=0.18\linewidth]{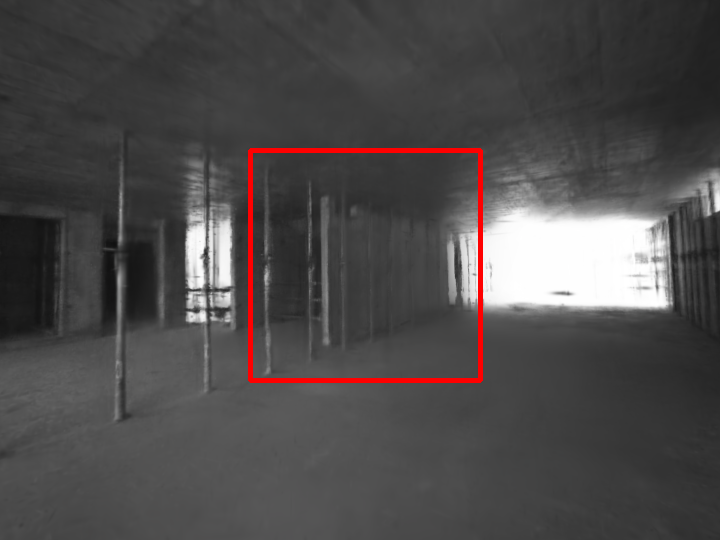} &
        \includegraphics[width=0.18\linewidth]{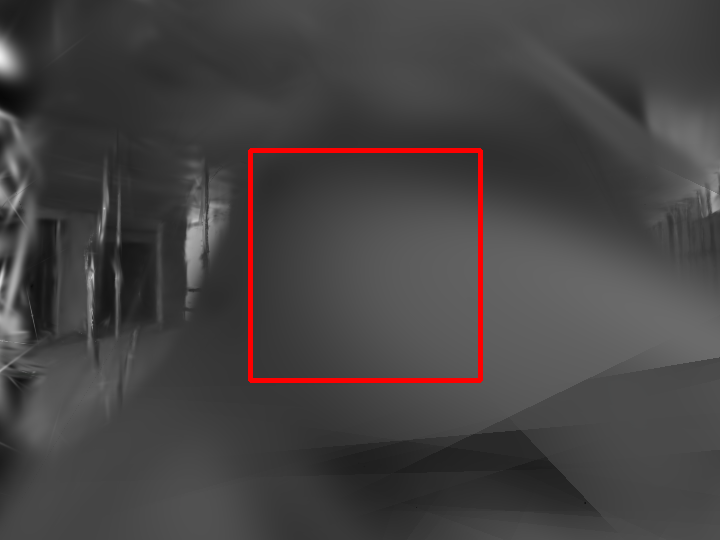}&
        \includegraphics[width=0.18\linewidth]{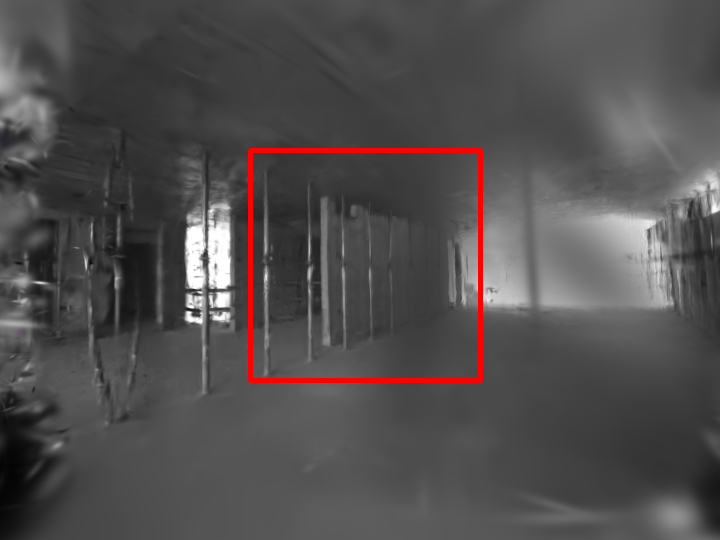}\\
        \includegraphics[width=0.18\linewidth]{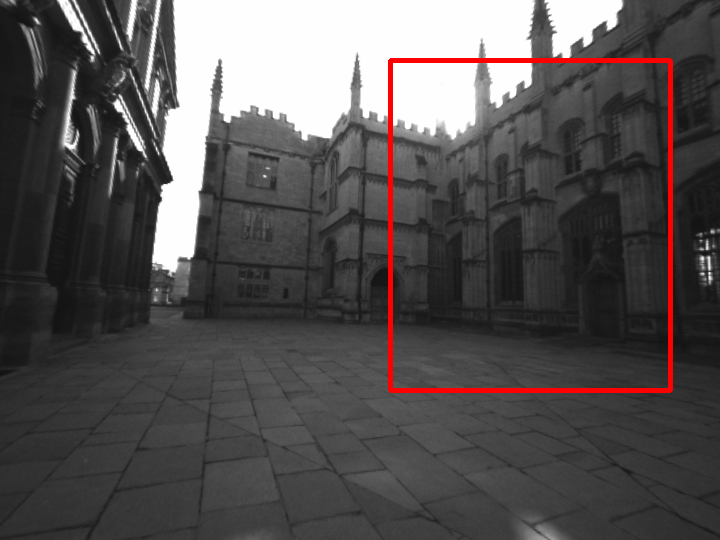} &
        \includegraphics[width=0.18\linewidth]{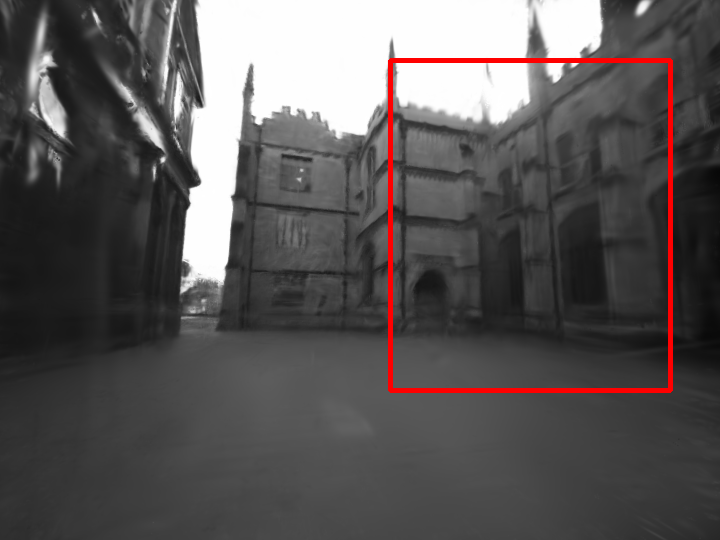} &
        \includegraphics[width=0.18\linewidth]{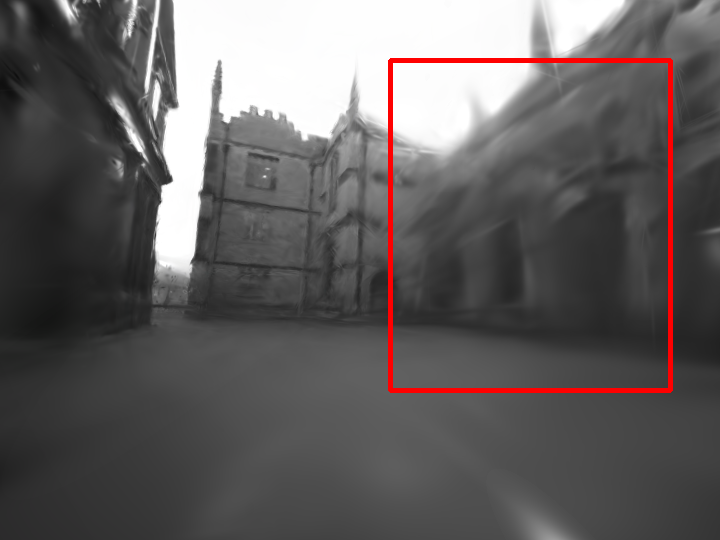} &
        \includegraphics[width=0.18\linewidth]{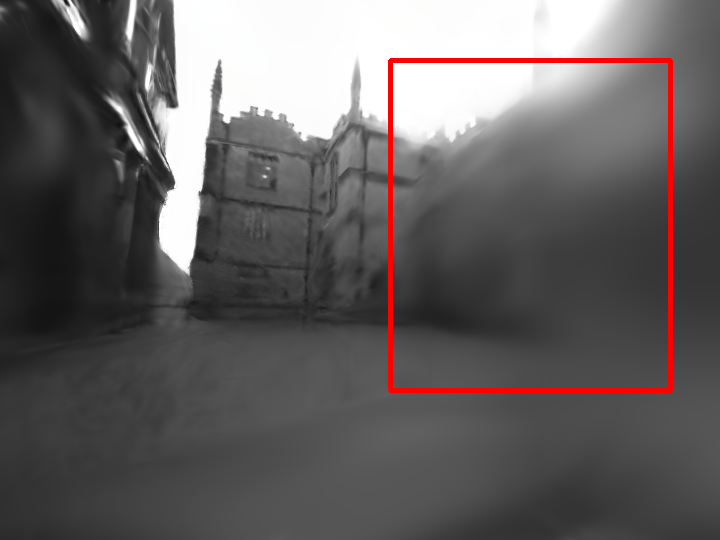}&
        \includegraphics[width=0.18\linewidth]{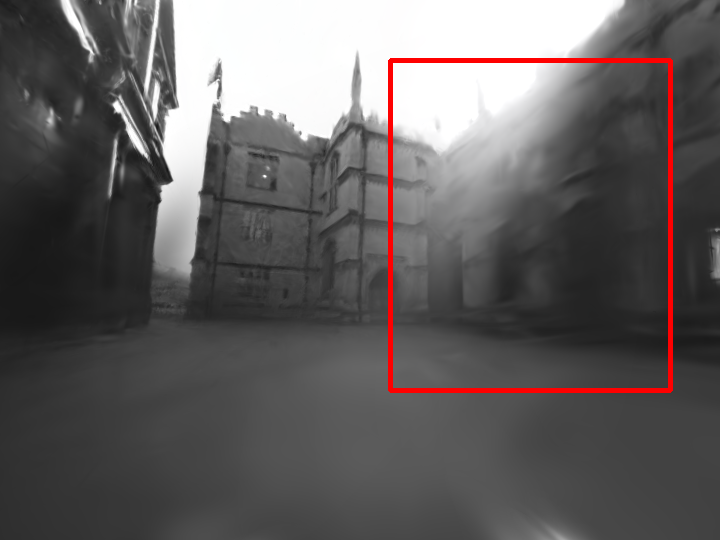}\\
    \end{tabular}
    \caption{Comparison of rendered images from different Gaussian Splatting methods on HILTI22 Dataset. Red boxes highlight key regions where our method preserves fine geometric and photometric details more accurately than competing models.}
    \label{fig:HILTI22_figures}
\end{figure*}

\begin{figure*}[h]
    \centering
    \renewcommand{\arraystretch}{0.8} 
    \setlength{\tabcolsep}{1pt} 
    \captionsetup{belowskip= 0pt} 
    \begin{tabular}{ccccc}
         \textbf{Ground Truth} &\textbf{Structured-Li-GS (Ours)} & \textbf{Scaffold-gs}& \textbf{2DGS} & \textbf{3DGS} \\
        \includegraphics[width=0.18\linewidth]{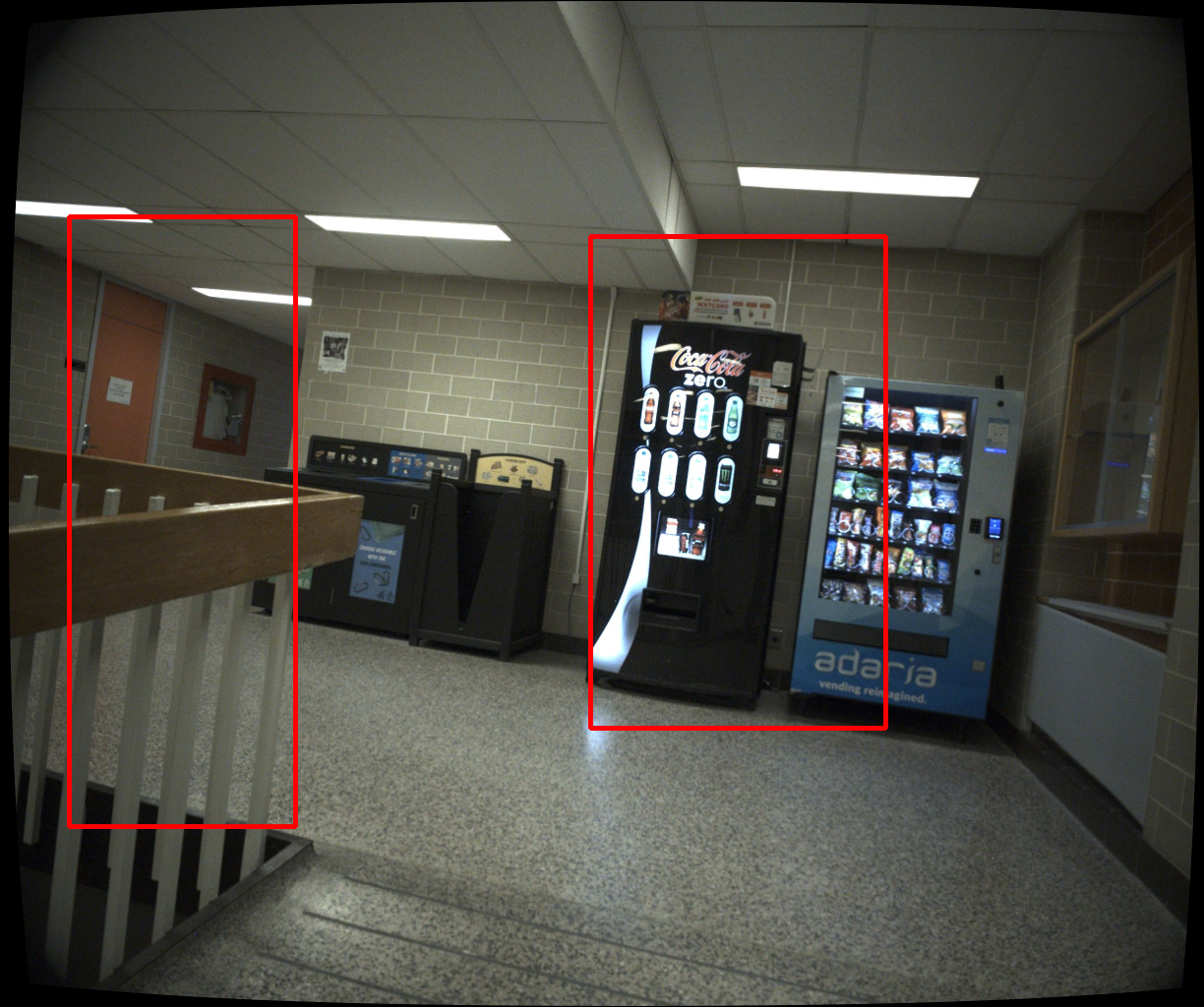} &
        \includegraphics[width=0.18\linewidth]{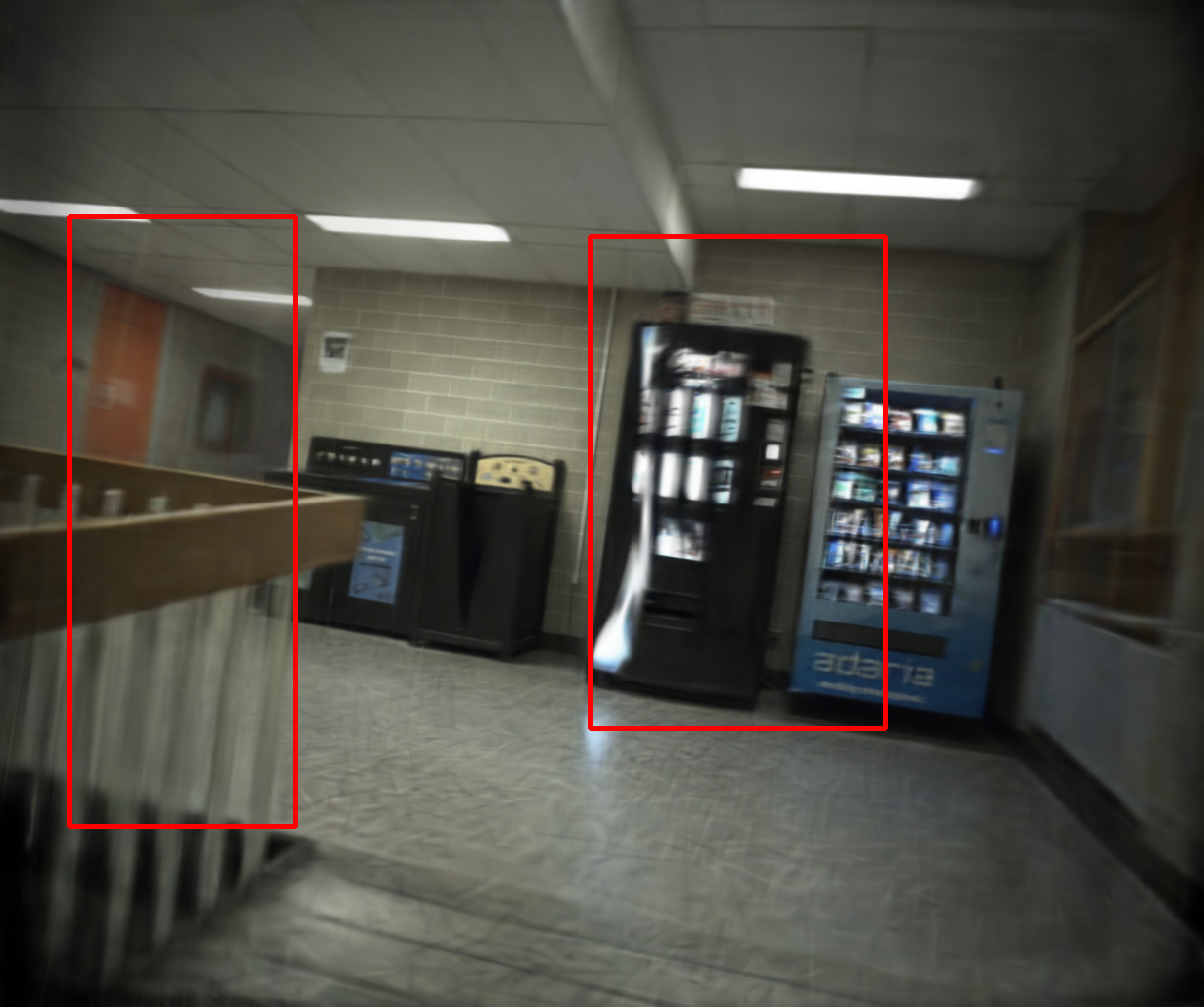}&
        \includegraphics[width=0.18\linewidth]{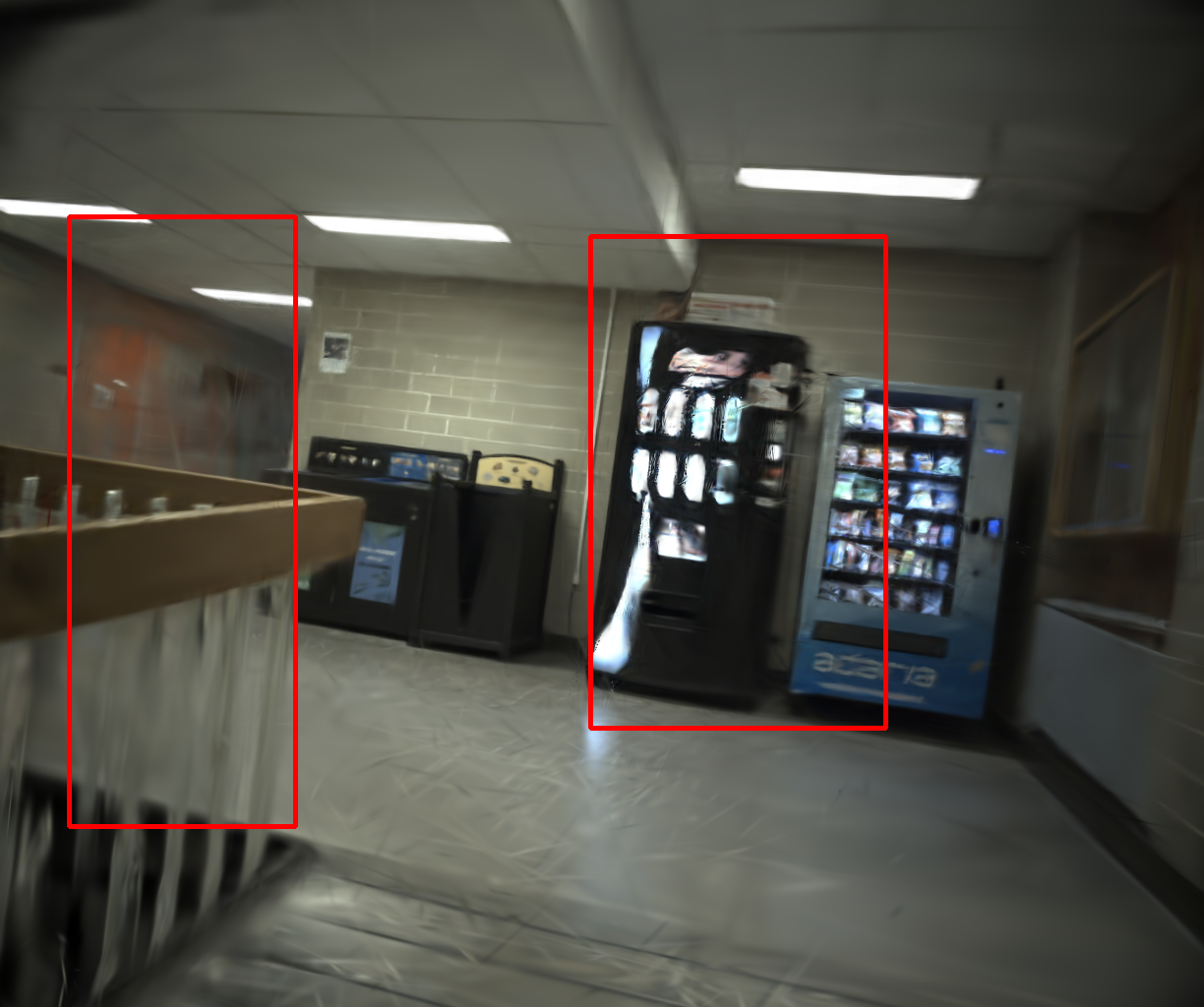} &
        \includegraphics[width=0.18\linewidth]{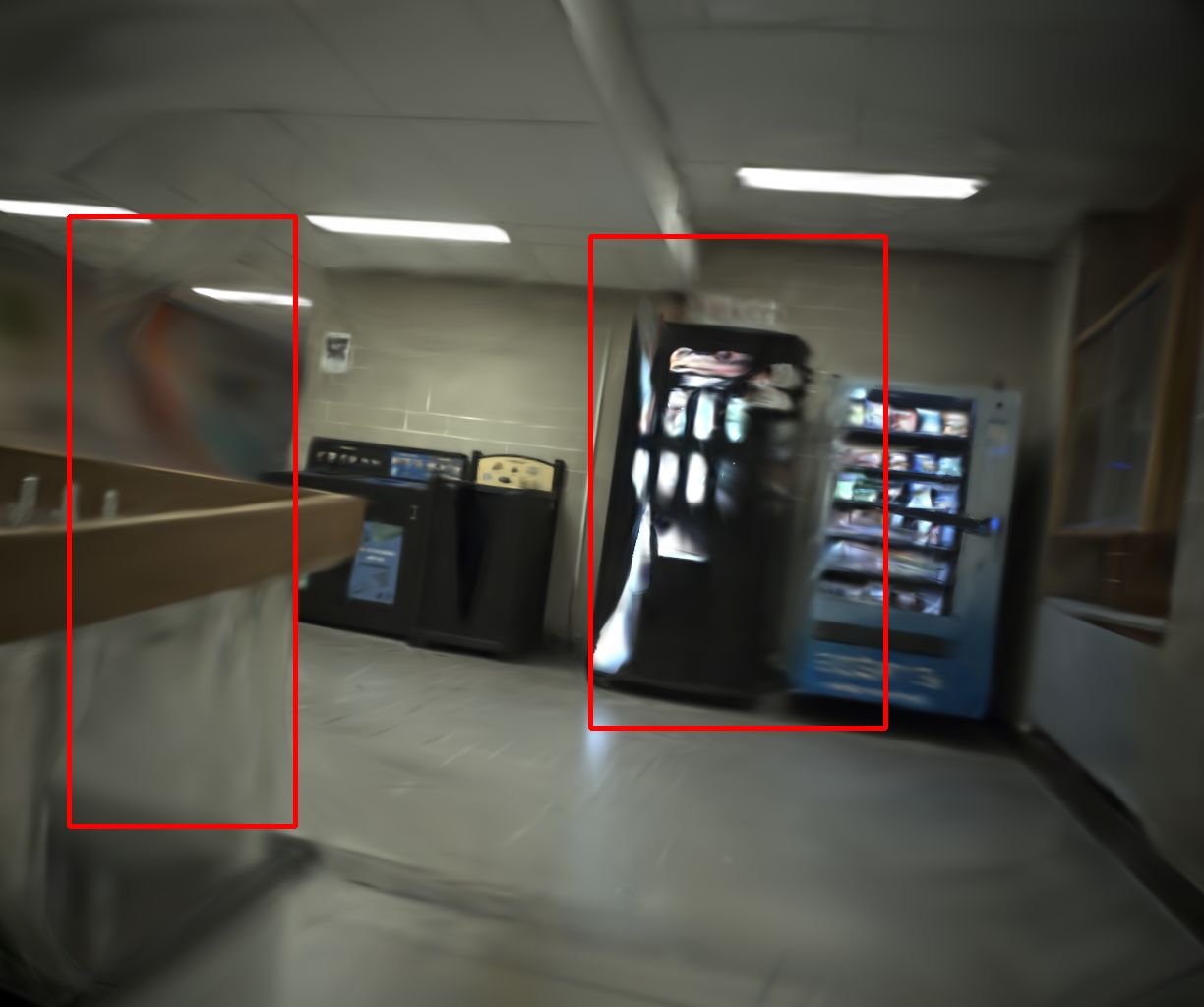}&
        \includegraphics[width=0.18\linewidth]{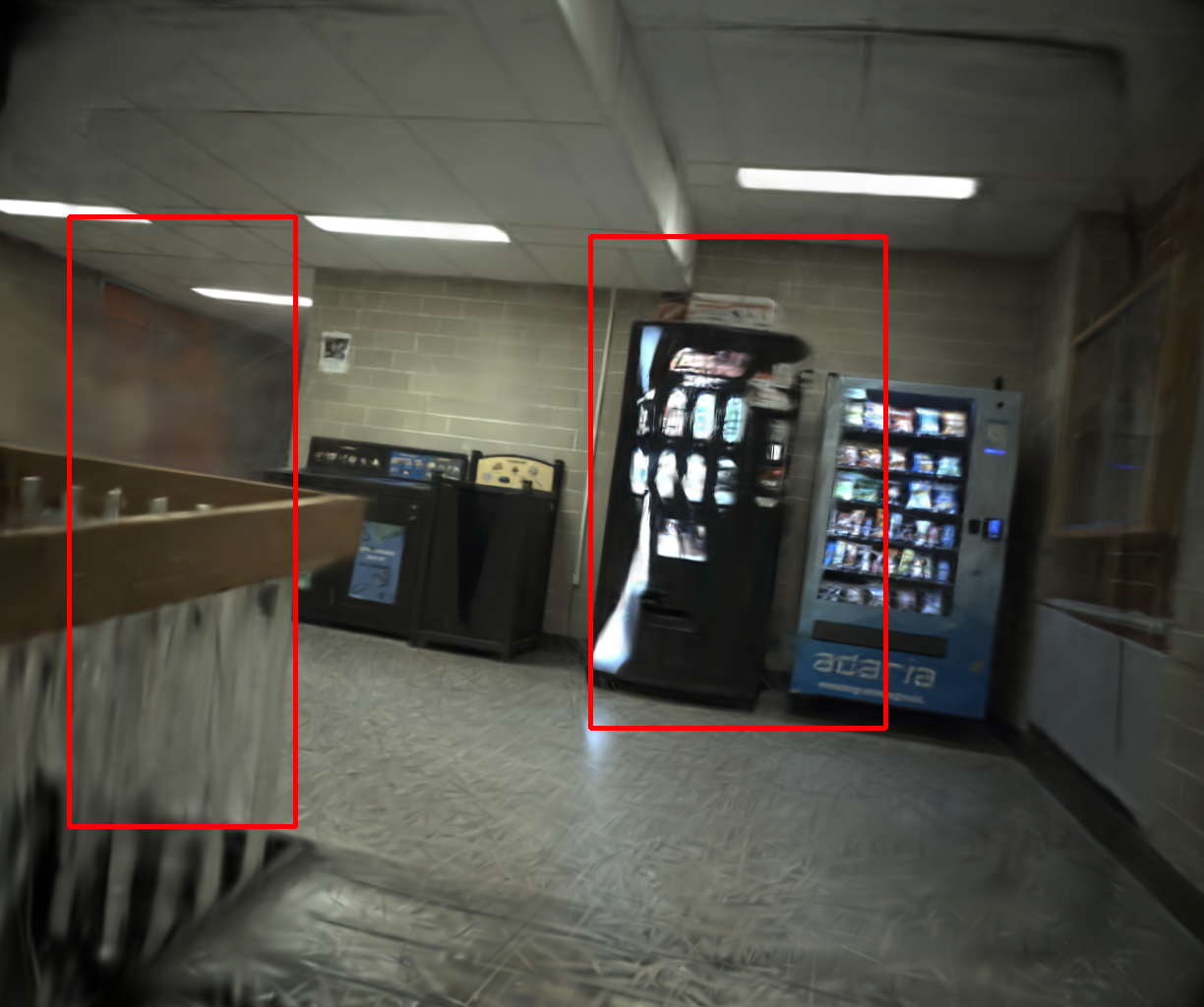}\\
        \includegraphics[width=0.18\linewidth]{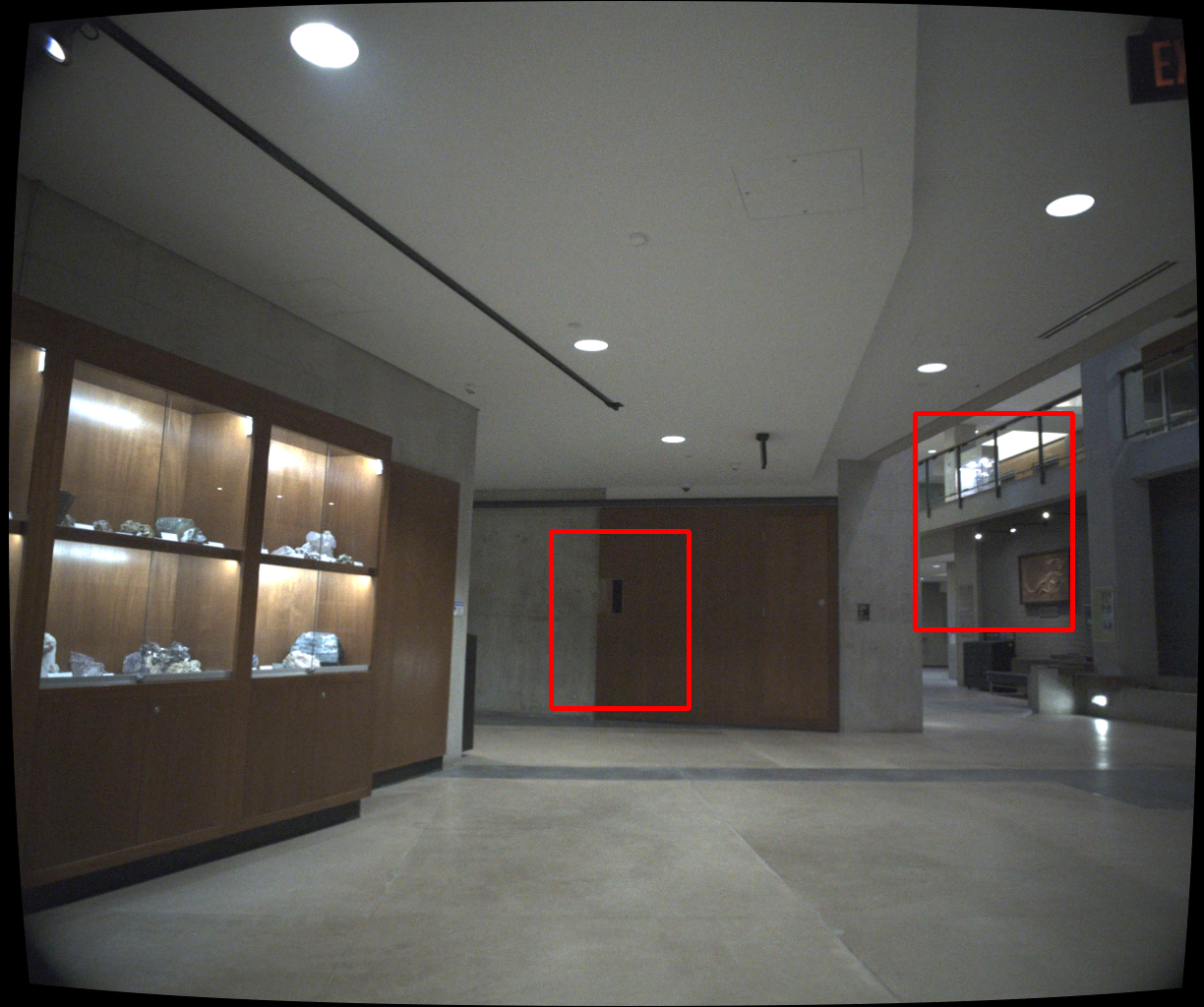} &
        \includegraphics[width=0.18\linewidth]{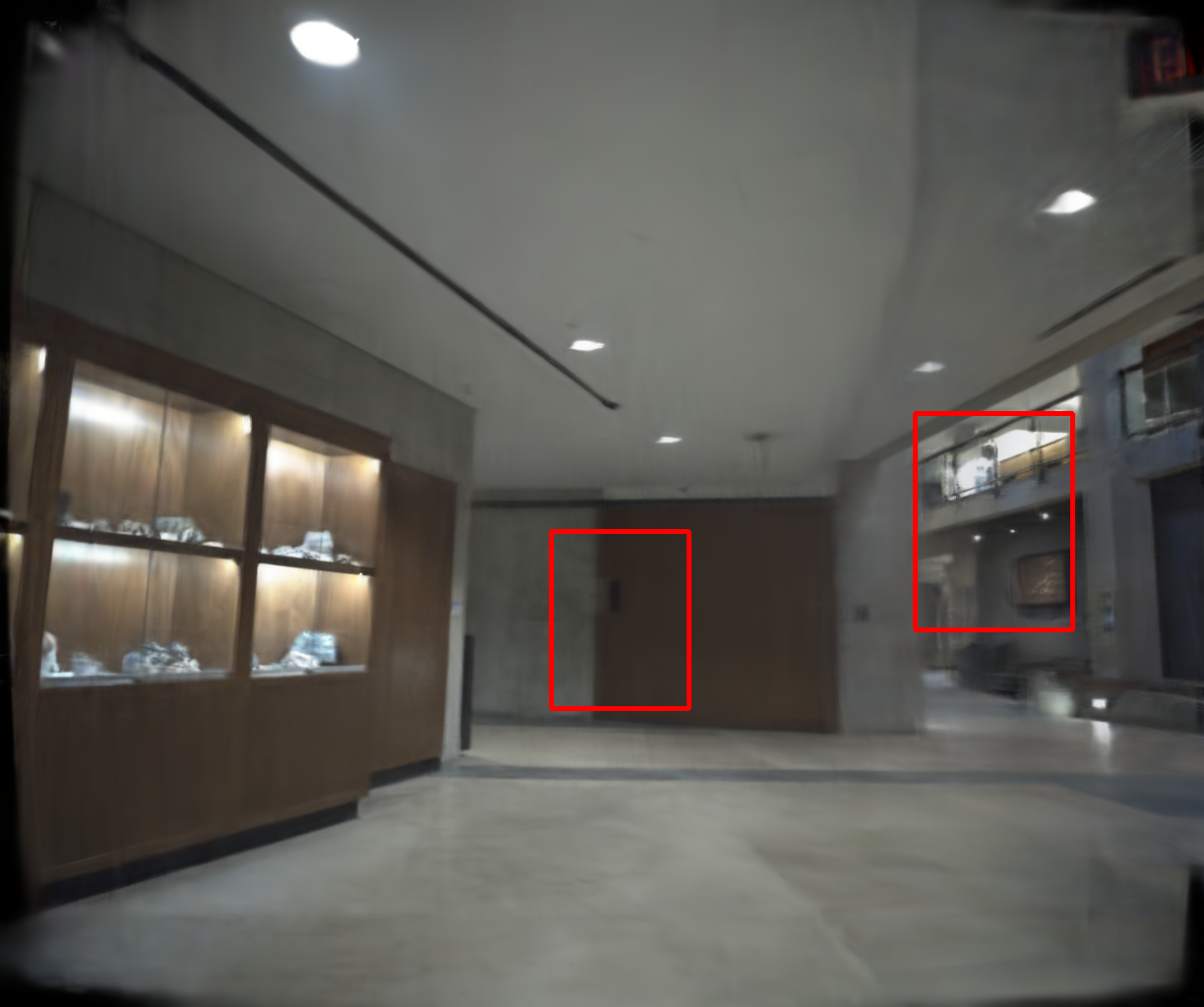}&
        \includegraphics[width=0.18\linewidth]{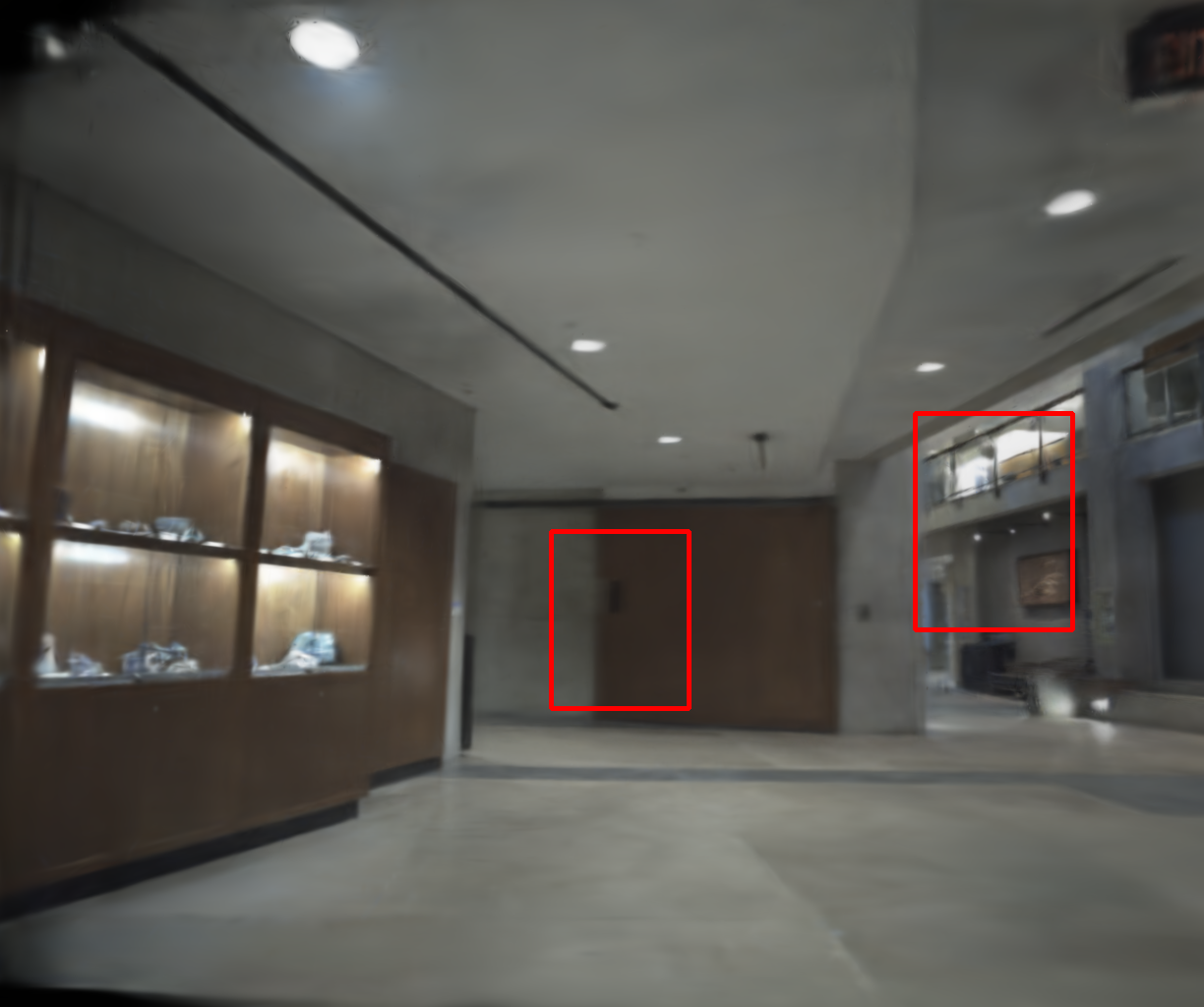} &
        \includegraphics[width=0.18\linewidth]{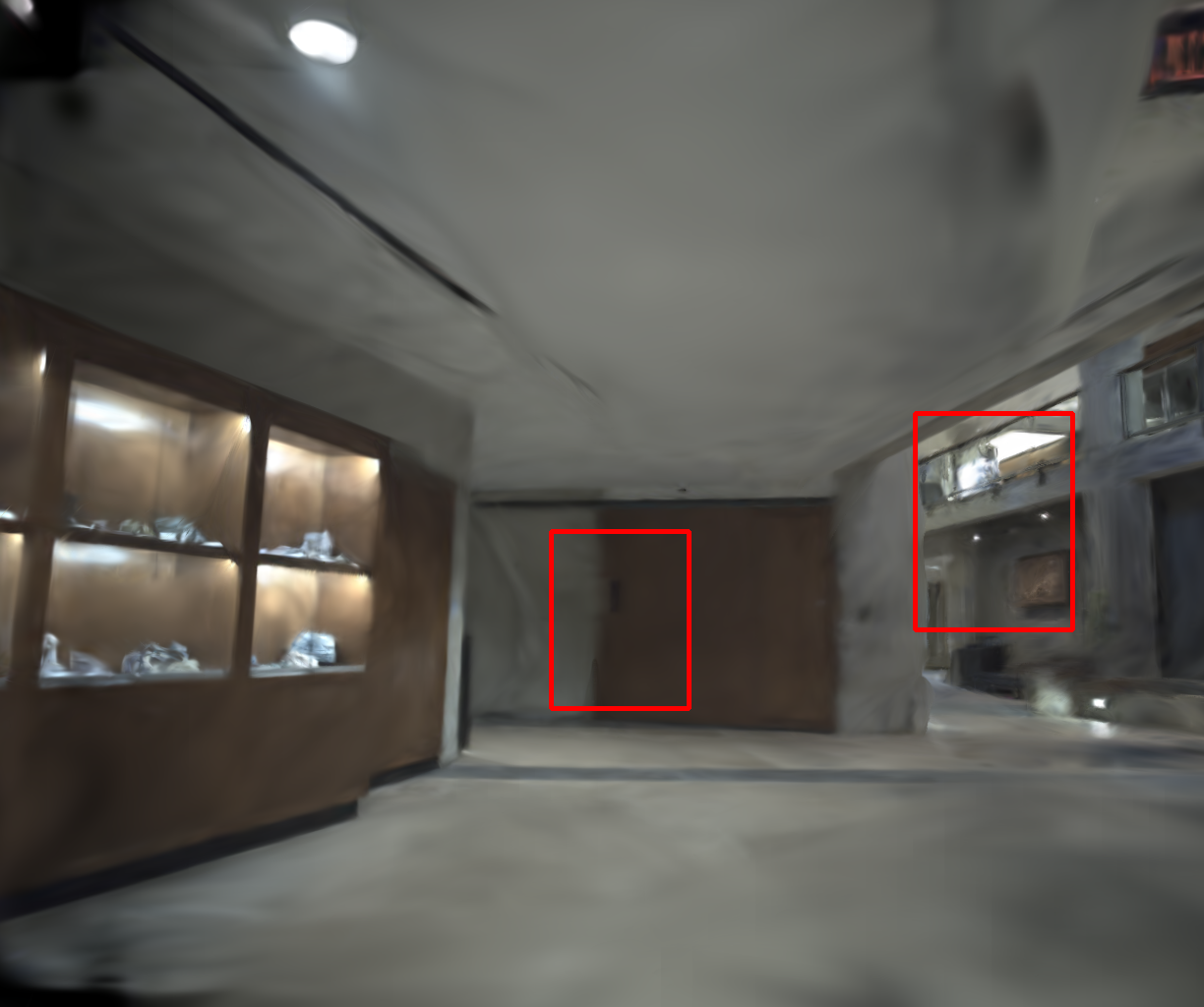}&
        \includegraphics[width=0.18\linewidth]{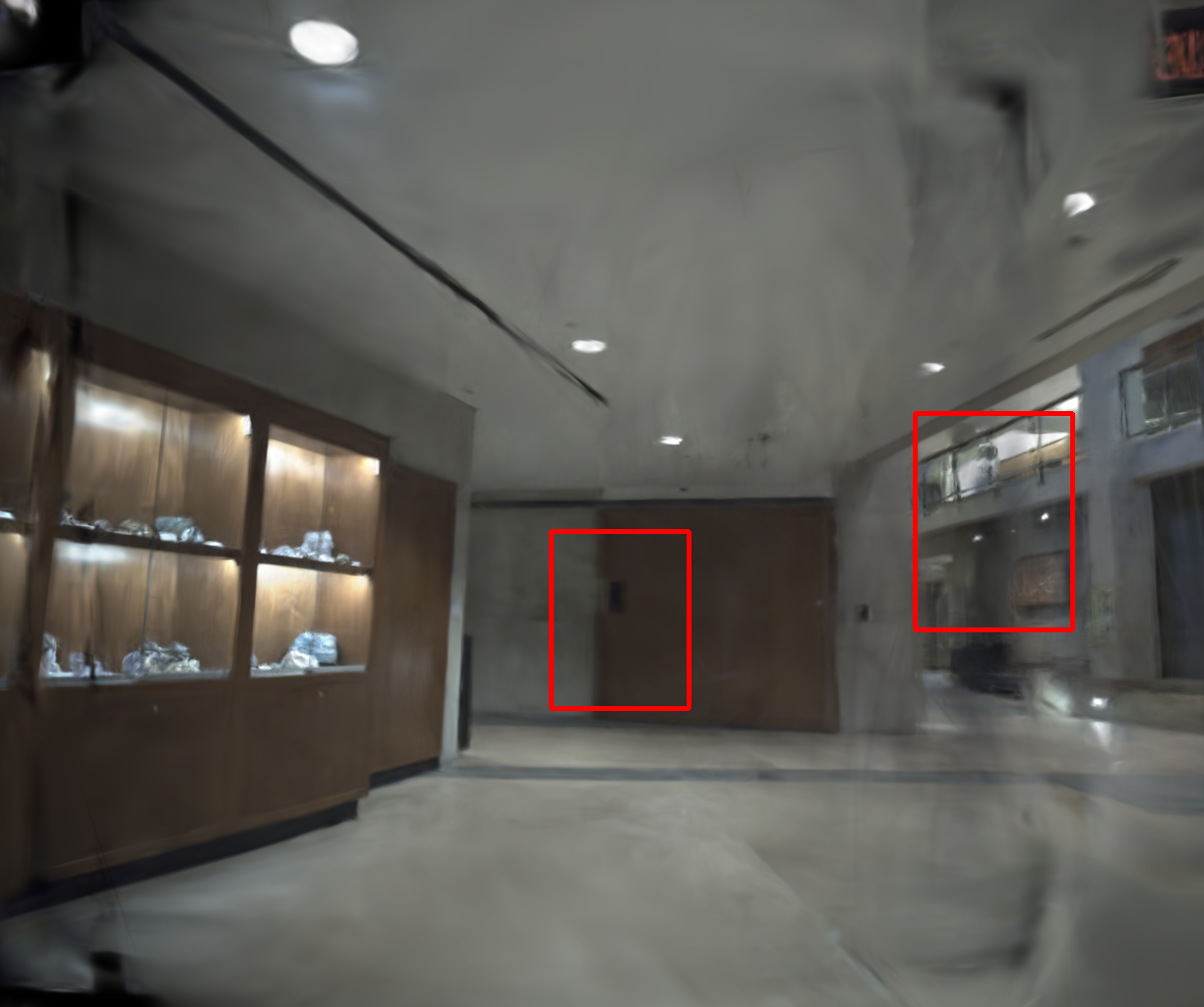}\\
    \end{tabular}
    \caption{Comparison of rendered images from different Gaussian Splatting methods on Custom Dataset. Red boxes highlight key regions where our method preserves fine geometric and photometric details more accurately than competing models.}
    \label{fig:custom_dataset_figures}
\end{figure*}

\subsection{Benchmark Comparation}

\subsubsection{FASTLIVO2 Dataset}
\text{                       } \\
Table~\ref{tab:comparison_table} and Fig.~\ref{fig:comparison_figures} together present a comprehensive evaluation of rendering quality across five challenging sequences from the FAST-LIVO2 dataset. Overall, our method demonstrates consistently strong performance, achieving top-tier results across nearly all metrics and scenes. In the CBD2 sequence, characterized by large-scale outdoor structures with sparse features—our approach attains the highest PSNR (22.82) and lowest LPIPS (0.389), while ranking second in SSIM (0.760). The corresponding qualitative renderings in Fig.~\ref{fig:comparison_figures} further highlight sharper geometric detail and fewer visual artifacts, confirming that our LiDAR-guided structured modeling better preserve both geometry and perceptual fidelity compared to baselines.

In sequences with richer texture and consistent viewpoint overlap, such as Retail Street and SYSU, our method continues to perform competitively or superiorly. For Retail Street, our method ranks second in PSNR and SSIM and maintains a comparable LPIPS, performing nearly identically to 3D-GS despite the challenges of long-distance viewing and dynamic lighting. As shown in four cases in Fig.~\ref{fig:comparison_figures}, our reconstructions retain color consistency and surface continuity where competing methods exhibit blurring or depth misalignment. Notably, on the SYSU sequence, our method achieves the best results across all three metrics (PSNR = 24.79, SSIM = 0.733, LPIPS = 0.362), reflecting robustness in cluttered, corridor-like environments where structured geometric priors provide greater stability.

The Main Building sequence presents one of the most complex reconstruction scenarios, involving mixed indoor–outdoor transitions and repetitive facade patterns. Even in this difficult case, our method again surpasses all baselines, achieving the best scores in every metric (PSNR = 26.15, LPIPS = 0.288). Visual comparisons in Fig.~\ref{fig:comparison_figures} demonstrate cleaner structural boundaries and reduced texture distortion. In sequences where our method is not the absolute leader in SSIM—such as CBD2—the gap is marginal and primarily caused by mild smoothing effects from our geometry-regularized representation. Nonetheless, our consistently lower LPIPS across most sequences indicates superior perceptual realism, aligning more closely with human visual quality.

Overall, the quantitative and qualitative results confirm that our Structured-Li-GS framework generalizes effectively across diverse environments, producing high-fidelity reconstructions with strong robustness to texture sparsity, lighting variation, and viewpoint diversity in the FAST-LIVO2 dataset.

\subsubsection{HILTI22 Dataset}
\text{                          }\\
Table~\ref{tab:comp_hilti_custom}, together with the visual comparisons in Fig.~\ref{fig:HILTI22_figures}, presents the rendering performance of different methods on the HILTI22 dataset. Overall, our Structured-Li-GS achieves strong and consistent results across all three sequences, ranking first or second in nearly every metric. In the Exp04 Construction sequence, featuring a mix of indoor and outdoor elements—our method achieves the highest PSNR (23.96) and lowest LPIPS (0.396), while slightly trailing Scaffold-GS in SSIM (0.839 vs. 0.841). The qualitative renderings in first row of Fig.~\ref{fig:HILTI22_figures} demonstrate sharper object boundaries and more stable color consistency, highlighting the advantage of our LiDAR-guided structured representation for mixed-structure environments.

In the Exp21 Outside Building sequence, which involves large open areas and repetitive textures, our method again achieves the best PSNR (24.31) and LPIPS (0.312), while maintaining competitive SSIM (0.832). As shown in the second row of Fig.~\ref{fig:HILTI22_figures}, our reconstructions retain global alignment and fine-grained texture details where other Gaussian-based methods exhibit geometric drift or surface smoothing. These results confirm that Structured-Li-GS maintains strong robustness in large-scale outdoor environments with sparse texture and wide-baseline viewpoints.

For the Exp14 Basement sequence, our method ranks second in PSNR (38.43) and LPIPS (0.169), slightly behind 3D-GS. The small gap can be attributed to the scene’s regular structure and limited viewpoint diversity, which favor dense reconstruction assumptions. Nevertheless, as illustrated in Fig.~\ref{fig:HILTI22_figures}, the visual quality remains comparable, demonstrating that Structured-Li-GS effectively preserves detail and perceptual realism even in texture-sparse indoor environments. Overall, these results show that our framework generalizes well across varying scene types, delivering high-fidelity reconstructions and robust perceptual quality on the public HILTI22 dataset.

\subsection{Custom Dataset Comparison}
On the Custom Dataset, which includes the Corride and EIT Building sequences, our Structured-Li-GS consistently demonstrates superior performance across all evaluation metrics. As summarized in Table~\ref{tab:comp_hilti_custom}, the method achieves either the best or second-best scores in every case, confirming its strong adaptability to diverse indoor and corridor-like environments captured by our custom handheld scanner.

In the Corride sequence, our model achieves the highest PSNR (25.58) and lowest LPIPS (0.457), indicating sharpness and perceptual fidelity, while ranking second in SSIM (0.773), only slightly below Scaffold-GS (0.779). Qualitative results in the first row of Fig.~\ref{fig:custom_dataset_figures} illustrate that our reconstructions preserve consistent texture continuity along elongated corridors, where competing methods often exhibit blur or color drift. Similarly, in the EIT Building sequence, our approach achieves the best scores across all three metrics—PSNR (23.97), SSIM (0.770), and LPIPS (0.478), demonstrating strong robustness in cluttered indoor scenes with challenging illumination and geometric variation. The second row of Fig.~\ref{fig:custom_dataset_figures} further highlights the superior visual quality of our rendered results.

Compared with all baselines, our method achieves the most balanced performance across both sequences, effectively managing visual complexity and spatial variation. The consistently low LPIPS scores highlight the perceptual advantage of our LiDAR-guided structured representation, enabling geometrically coherent and visually faithful reconstructions even in real-world, sensor-collected data.

\subsection{Number of Gaussians Comparison}

\begin{table}[]
\centering
\caption{Comparison of the number of Gaussians used in the trained models on the CBD2 sequence of the FAST-LIVO2 dataset. The lowest value indicates higher efficiency. The best result is highlighted in \textbf{bold}, and the second-best is \underline{underlined}.}
\label{tab:number_comparison}
\begin{tabular}{lcc}
\toprule
\text{Method}  & \text{\# of Gauss.} $\downarrow$  \\
\midrule
    \text{3D-GS} & 1,893,221  \\
    \text{2D-GS}  & {929,169} \\
    \text{Scoffold-GS} & \underline{521,782} \\
    \text{LetsGo} & 1,005,452  \\
    \text{AtomGS}  &  2,005,056 \\
    \text{Structured-Li-GS (Ours)}  & \textbf{356,885}  \\
\bottomrule
\end{tabular}
\end{table}

Table~\ref{tab:number_comparison} presents a comparison of the number of Gaussian primitives used by different Gaussian Splatting (GS) variants on the CBD2 sequence of the FAST-LIVO2 dataset. The number of Gaussians directly reflects the model’s compactness and computational efficiency, as fewer Gaussians generally indicate reduced memory and rendering costs while maintaining reconstruction fidelity.

Among all methods, our proposed \text{Structured-Li-GS} achieves the most compact representation, requiring only \textbf{356,885} Gaussians—approximately one-third of those used by Scaffold-GS and nearly five times fewer than standard 3D-GS. This demonstrates the effectiveness of our LiDAR-guided initialization and structured training strategy, which leverages dense geometry priors to improve scene coverage without relying on excessive Gaussian density. In contrast, methods such as AtomGS and 3D-GS tend to overfit by allocating redundant Gaussians in textureless or planar regions, leading to unnecessarily large model sizes.

These results confirm that our framework achieves a more efficient balance between reconstruction quality and model compactness, providing a scalable solution for large-scale, high-fidelity 3D scene representation.

\begin{table}[]
\centering
\caption{Comparative Results of Ablation Study from CBD2 sequence in FASTLIVO2 Dataset. We report PSNR ($\uparrow$), LPIPS ($\downarrow$), and SSIM ($\uparrow$) for models. The best performance for each metric is highlighted in \textbf{bold}, and the second-best is \underline{underlined}.}
\label{tab:ablation_loss}
\begin{tabular}{lccc}
\toprule
\text{Loss Configuration} & \text{PSNR} & \text{SSIM}& \text{LPIPS} \\
\midrule
w/o flatten loss & 22.72 &  0.7574 & 0.3942  \\
w/o offset loss & \underline{22.77} &  0.7568 & 0.3969  \\
w/o depth loss & 22.75 &  \underline{0.7584} & 0.3927  \\
w/o normal loss & {22.71} & 0.7576 & \underline{0.3926} \\
w/ all losses & \textbf{22.82} & \textbf{0.7601} & \textbf{0.3896}  \\
\bottomrule
\end{tabular}
\end{table}

\subsection{Ablation Study}
We conducted an ablation study to evaluate the impact of different individual loss on reconstruction accuracy and rendering quality, with results shown in Table~\ref{tab:ablation_loss}. The experiments also used a voxel size of 0.0065m for point cloud downsampling and 0.005m for Gaussian voxels during training, with all metrics reported on the test dataset as same as the comparative study.

Structured-Li-GS with all loss functions outperformed all ablated versions, achieving the highest PSNR (22.82 dB), SSIM (0.7601), and lowest LPIPS (0.3896), demonstrating the effectiveness of the proposed comprehensive loss function. Removing individual loss components, such as the flatten loss, offset loss, depth loss, absolute normal loss, and normal consistency loss, results in slight degradations across all metrics. For instance, excluding the flatten loss reduces PSNR to 22.72 dB and SSIM to 0.7574, while removing the offset loss lowers PSNR to 22.77 dB. The model without any loss function performs the worst, with a PSNR of 22.07 dB and SSIM of 0.7406, emphasizing the importance of 
incorporating loss functions for effective learning. Overall, the results of the ablation study highlights that  to reconstruct the high-quality 3DGS model.

\section{Conclusion}
This paper proposes Structured-Li-GS, a structured and efficient 3D reconstruction approach that integrates LiDAR data and 3DGS. This integration leverages LiDAR measurements for anchor-based, normal-assisted Gaussian initialization, ensuring high rendering quality with a controlled model size by maintaining a fixed number of Gaussians. To enhance surface fitting and rendering accuracy in unbounded and large-scale scenes, we incorporate a multi-regularization loss during optimization. Experimental results demonstrate that Structured-Li-GS achieves superior performance compared to state-of-the-art methods, offering a balanced trade-off between reconstruction quality and model size.

For future work, we will extend our approach to street-level imagery with LiDAR, targeting large-scale outdoor 3DGS. To improve robustness, we plan to strengthen data preprocessing, particularly surface normal estimation and depth image generation. For scalability, we will integrate hierarchical 3D Gaussian representations \cite{hierarchicalgaussians24} to enable level-of-detail (LoD) training and rendering. Finally, we will incorporate mesh extraction to complete the end-to-end reconstruction pipeline.

\bibliography{cited_works}

\end{document}